\documentclass{article}

\usepackage{PRIMEarxiv}

\usepackage[utf8]{inputenc} 
\usepackage[T1]{fontenc}    
\usepackage[colorlinks=true, citecolor=blue]{hyperref} 
\usepackage{url}            
\usepackage{booktabs}       
\usepackage{amsfonts}       
\usepackage{nicefrac}       
\usepackage{microtype}      
\usepackage{lipsum}
\usepackage{fancyhdr}       
\usepackage{graphicx}       
\graphicspath{{media/}}     
\usepackage{natbib} 

\usepackage{microtype}
\usepackage{graphicx}
\usepackage{subfigure}
\usepackage{booktabs}
\usepackage{xcolor}
\usepackage{caption}
\usepackage{subcaption}
\usepackage{wrapfig}
\usepackage{graphics}
\usepackage{enumitem}
\usepackage{amsmath}
\usepackage{amssymb}
\usepackage{mathtools}
\usepackage{amsthm}
\usepackage[skip=10pt plus1pt, indent=0pt]{parskip}
\let\svthefootnote\thefootnote
\newcommand\freefootnote[1]{%
  \let\thefootnote\relax%
  \footnotetext{#1}%
  \let\thefootnote\svthefootnote%
}

\providecommand{\norm}[1]{\lVert#1\rVert}

\providecommand{\vx}{\vec{x}}

\providecommand{\vz}{\vec{z}}

\providecommand{\vw}{\vec{w}}

\renewcommand{\vec}[1]{\ensuremath{\boldsymbol{#1}}}

\newcommand{\gauss}[2]{\mathcal{N}\left( #1,#2 \right)} 
\providecommand{\loss}{\mathcal{L}}
\newcommand*\diff{\mathop{}\!\mathrm{d}} 

\newcommand{\errbar}[2]{#1 {\scriptsize (#2)}}

\title{Distributional Dataset Distillation with Subtask Decomposition
}

\author{
  Tian Qin \\
  Harvard University \\
  Cambridge, MA\\
   \And
  Zhiwei Deng \\
  Google Research \\
  Mountain View, CA\\
  \And
  David Alvarez-Melis \\
  Harvard University \& MSR \\
  Cambridge, MA\\
}

\begin{document}
\maketitle

\begin{abstract}
What does a neural network learn when training from a task-specific dataset? Synthesizing this knowledge is the central idea behind Dataset Distillation, which recent work has shown can be used to compress large datasets into a small set of input-label pairs (\textit{prototypes}) that capture essential aspects of the original dataset. In this paper, we make the key observation that existing methods distilling into explicit prototypes are very often suboptimal, incurring in unexpected storage cost from distilled labels. In response, we propose \textit{Distributional Dataset Distillation} (D3), which encodes the data using minimal sufficient per-class statistics and paired with a decoder, we distill dataset into a compact distributional representation that is more memory-efficient compared to prototype-based methods. To scale up the process of learning these representations, we propose \textit{Federated distillation}, which decomposes the dataset into subsets, distills them in parallel using sub-task experts and then re-aggregates them. We thoroughly evaluate our algorithm on a three-dimensional metric and show that our method achieves state-of-the-art results on TinyImageNet and ImageNet-1K. Specifically, we outperform the prior art by $6.9\%$ on ImageNet-1K under the storage budget of 2 images per class.
\freefootnote{Correspondence to: Tian Qin $\langle$tqin@g.harvard.edu$\rangle$.}
\end{abstract}


\section{Introduction}
\label{intro}

Large datasets such as ImageNet \citep{deng2009large} can be used for a variety of purposes, ranging from image classification, single-object localization to generative tasks. If one only needs to accomplish one of those tasks, say image classification, can we synthesize only relevant information in the data and thus achieve compression? The goal of data distillation, first introduced by \citet{wang2018dataset}, is to answer this question: how to `condense' a dataset into a smaller (synthetic) counterpart, such that training on this distilled dataset achieves performance comparable to training on the the original dataset. Since its inception, this problem has garnered significant attention due to its obvious implications for data storage efficiency, faster model training, and democratization of large-scale model training. It also holds the promise of speeding up downstream applications such as neural architecture search, approximate nearest neighbor retrieval, and knowledge distillation, all of which often require data-hungry methods \citep{sachdeva2023data}. Moreover, data distillation has emerged as a promising approach for continual learning \citep{rosasco2021distilled} and differential privacy \citep{dong2022privacy}, often outperforming bespoke differentially-private data generators both in terms of performance and privacy, and allowing for private medical data sharing \citep{li2022compressed}.

\begin{figure*}[ht]
\begin{center}
\centerline{
\includegraphics[width=0.9\textwidth]{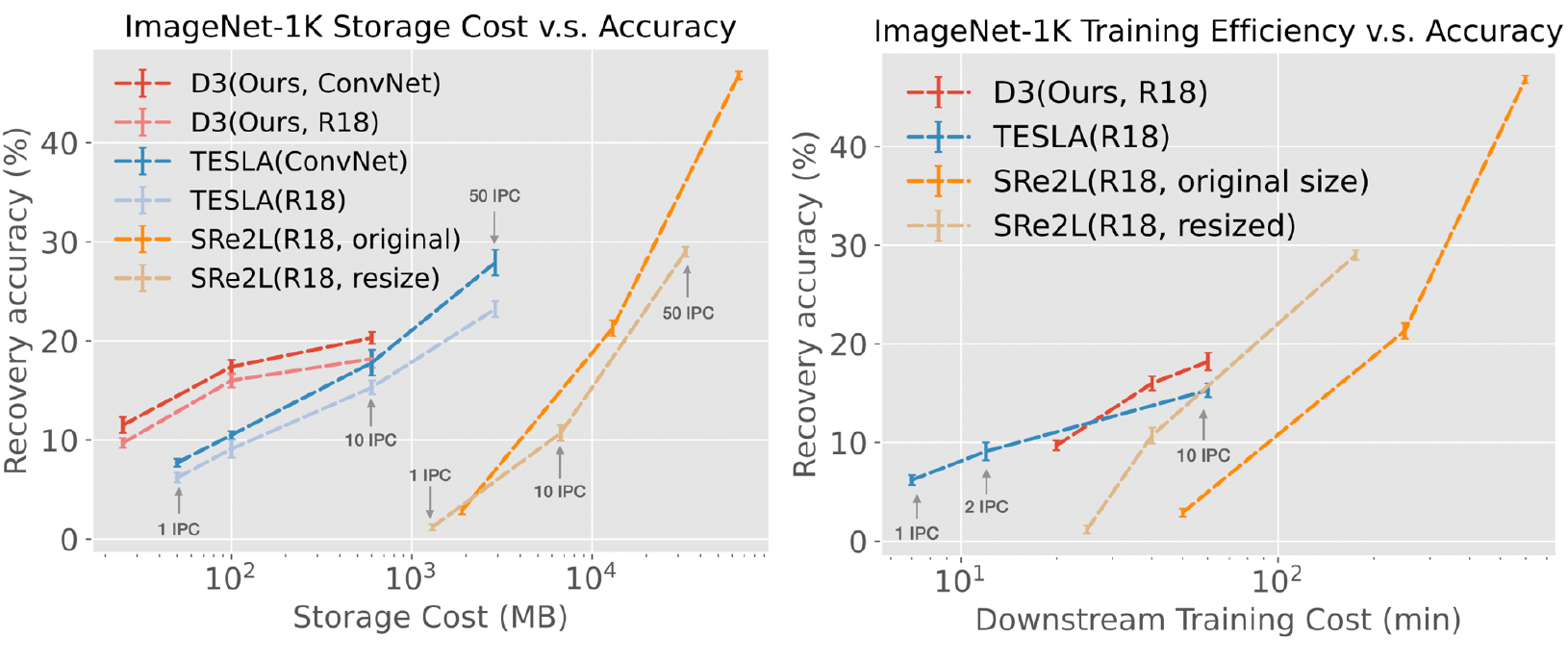}

}
\caption{\textbf{Three-dimensional evaluation on methods that scale to ImageNet-1K.} \textit{Left:} Recovery accuracy vs.~storage trade-off comparison for our (D3) and other methods on resized ($64\times64\times3$) ImageNet-1K. Our method achieves SOTA performance at small memory cost regime. \textit{Right:} Accuracy vs.~downstream task training cost on resized ImagNet-1K. }
\label{fig:pull}
\end{center}
\vspace{-0.25in}
\end{figure*}


Most current state-of-the-art data distillation methods produce \textit{synthetic protoypes}: a small subset of learnt (input, label) pairs that capture the most `salient' (in terms of their impact on classifier performance) aspects of the original dataset. These prototypes are often defined in the original input (e.g., image) space (\cite{wang2018dataset, wang2022cafe}). Recently, some work \citep{deng2022remember, zhao2022synthesizing,leehb2022dataset} propose to distill images into a latent space, and use a decoder to map latent codes back to the input (image) space.
Overall, dataset distillation methods have achieved remarkable success in producing much smaller datasets\,---typically measured in terms of Images (or Prototypes) Per Class (IPC)---\, with limited loss of downstream model performance. While early methods suffered from limited scalability, recent ones have managed to scale to large datasets like ImageNet-1K or even ImageNet-21K \citep{yin2023dataset_2, yin2023squeeze, liu2023dataset}. For example, SRe$^2$L \citep{liu2022datset} achieved a $\sim\!\!100\times$ IPC reduction on ImageNet-1K and recovered $\sim\!\!77\%$ of the classification accuracy \footnote{SRe$^2$L used ResNet18 as the teacher model, which achieved $69.8\%$ classification accuracy from full ImageNet-1K training. SRe2L's distilled data with 10 IPC achieved $46.2\%$ classification accuracy}.




Although encouraging, we will show that these results tell an incomplete story. When considering the \textit{total storage} (e.g., disk space used to store all necessary distillation outputs) and the runtime needed to train new models on the distilled data, the efficacy of these methods is much more subdued. Beyond the prototypes, some of these methods output other artifacts that are necessary for downstream use but whose memory footprint is rarely reported. These include soft labels (often multiple per prototype) and augmentation parameters used \citep{zhou2023dataset, yin2023dataset_2, yin2023squeeze}. The use of distilled labels are crucial (see Figure~\ref{fig:nosoft_compare}) but incur a storage cost that is not captured by IPC (Table~\ref{tab:storage_expanded}, Appendix \ref{appdx:fig1}). Once we take into account the storage cost of these artifacts, the true compression rate of such methods is much lower than implied by the IPC metric (Figure~\ref{fig:pull}, left). On the other hand, decoding/generation/augmentation procedures often translate into additional post-distillation training time (Figure~\ref{fig:pull}, right). In light of these observations, we argue that IPC as a metric of distillation is incomplete, and that the methods that have been developed to optimize it should be revisited with a more comprehensive set of evaluation metrics.






\begin{wrapfigure}[16]{r}{0.5\textwidth}
\begin{center}
\vspace{-0.1in}
\centerline{
\includegraphics[width=0.5\textwidth]{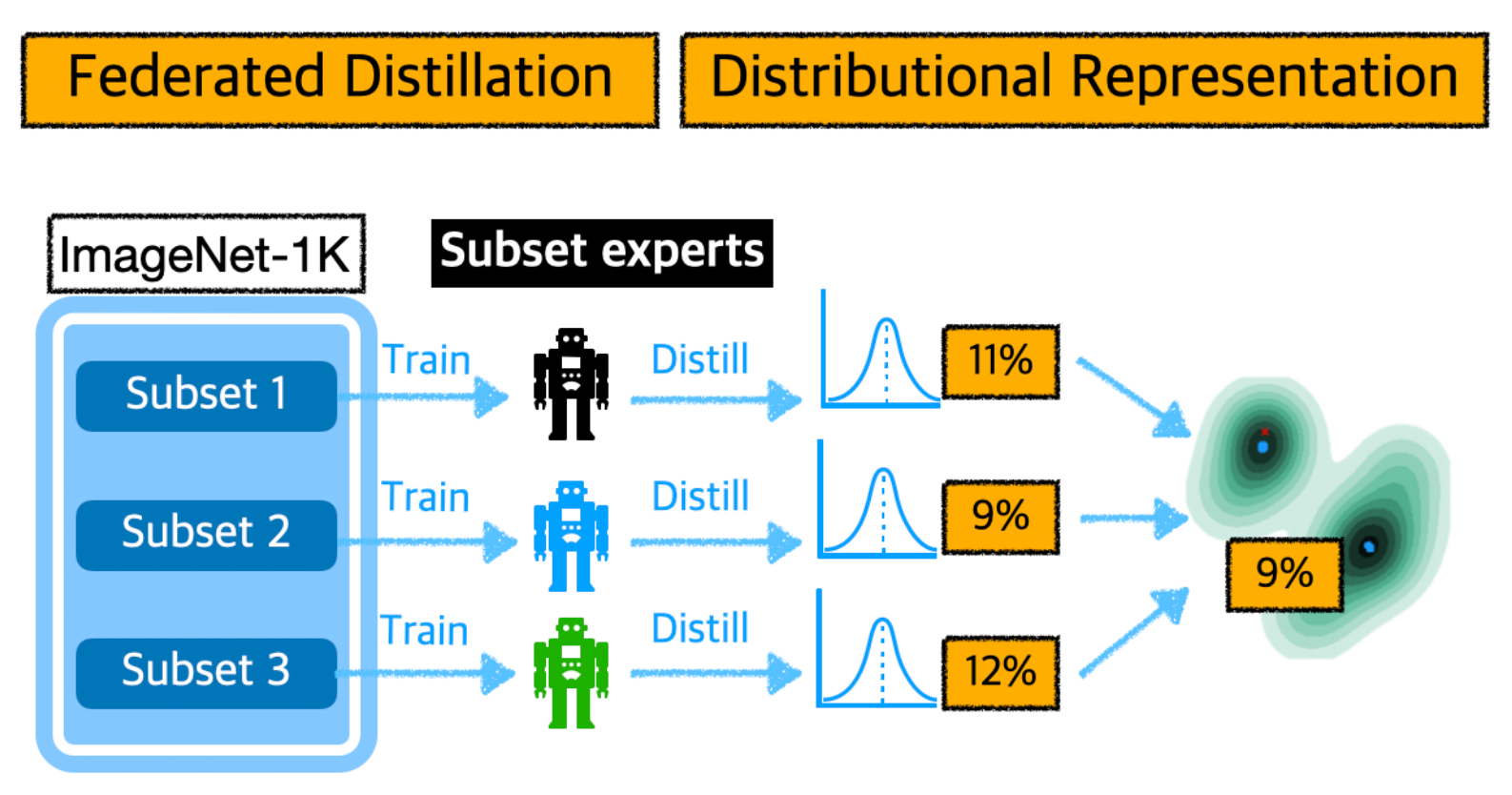}
}
\caption{\textbf{Illustration of Federated Distillation and Distributional Representation} We decompose large datasets into subtasks and distill each subset into distributions using locally trained experts. Distributions distilled  on subtasks generalize well to the full task.}
\label{fig:pull_demo}
\end{center}
\end{wrapfigure} 
In response to the above observation, we propose a new dataset distillation method with efficient compression properties, not in terms of dataset cardinality (i.e., IPC), but directly in terms of storage size and downstream model training time. 
Distilling into the latent space not only allows a more compact representation of the data by sharing inter-class mutual information in decoder parameters, but also offers finer-grained control on compression than working directly with prototypes (e.g., by varying number of latent codes per class, latent dimensions and the decoder size). We challenge the conventional approach of distilling into a finite set of (latent) prototypes and propose to cast the problem into a \textit{distributional} one: finding a synthetic probability distribution which can be sampled to produce training data for downstream tasks. This Distributional Dataset Distillation (D3) approach yields an efficient representation of distilled data without incurring much additional computation costs on downstream tasks.

To scale our method efficiently to large datasets such as ImageNet-1K, we propose a simple-yet-effective \textit{federated distillation} scheme that parallelizes the distillation process (Figure \ref{fig:pull_demo}). Instead of directly distilling the entire dataset, we divide the full classification tasks into subtasks, where each task only classifies a subset of all classes. Data distillation is performed on subtasks, using local experts trained on subtasks. We then aggregate the locally distilled datasets to form the distilled data for the full task. We show that data distilled on subtasks generalize well to the full task, which ensures the good performance of our federated distillation process. Using the distributional representation and federated distillation, we achieve SOTA performance on ImageNet-1K as measured by storage cost. 

Our contributions can be summarized as follows: 
\begin{itemize}
[itemsep=2pt,labelindent=0pt,topsep=0pt,parsep=0pt,partopsep=2pt, align=left]
    \item We show that state-of-the-art prototype-based data distillation methods yield unexpectedly high storage costs and post-distillation training times, an under-reported phenomenon that is not captured by commonly-used compression metrics (e.g., IPC, the \textit{number} of distilled items per class). The large storage cost and training time could hinder the usability of these methods in practice. 
    \item We propose a novel distillation framework with smaller memory footprint that distills datasets into \textit{distributions}, extending recent methods that distill into a latent space to now operate on (latent) distributions. We show this method matches or outperforms state-of-the-art distillation methods in terms of prediction accuracy on various datasets (e.g., TinyImageNet, ImageNet-1K), with smaller storage costs. 
    \item We propose a simple-yet-effective federated distillation strategy that allows distillation training process to be parallelized, and which has general applicability beyond our specific method. 
    \item In response to our observations above, we propose new evaluation protocols and metrics for dataset distillation methods that more accurately characterize the extent of `distillation', and compare existing work and our work along these axes. \footnote{Code for all experiments is available here: \url{https://github.com/sunnytqin/D3}}
    

\end{itemize}

\section{Related Work}
\label{sec:related}


We focus our discussion of prior work on the lines that are most closely related to ours, but note that methods with similar goals have been developed in the context of statistical sample compression \citep{winter2002compression, dwivedi2021kernel} and core-set selection \citep{mirzasoleiman2020coresets, zhou2022probabilistic}. 

\textbf{Optimization Methods}
\citet{wang2018dataset} originally approached the distillation problem as a bi-level optimization task, which is computationally intensive. To tackle the computation challenge, many work has proposed proxy training objectives to simplify the distillation process. \citet{nguyen2020dataset} leveraged NTK-based algorithms to solve the inner optimization in closed form. \citet{zhao2020dataset} proposed gradient matching to avoid the unrolling of the inner-loop and make the distillation process more efficient. 
Further improvements on single-iteration gradient matching also include \citep{leehb2022dataset, zhao2021dataset}. 
Matching training trajectories (MTT) was proposed by \citep{cazenavette2023generalizing}, claiming that matching long-range training dynamics provides further improvements on single-iteration gradient matching. \citet{cui2022scaling} proposed TESLA as a scalable alternative to the original MTT method. \citet{du2022minimizing} proposed a variant that uses ``flat" trajectory matching to further improve trajectory-matching based methods. Distribution Matching (DM), proposed by \citep{zhao2023dataset}, seek to minimize the Maximum Mean Discrepancy (MMD) between original and distilled dataset samples. Further refinement on the method includes \citet{wang2022cafe, zhou2023dataset}. Specifically, Neural Feature Regression with Pooling (FRePo) \citep{zhou2023dataset} addressed the memory-concern with a pooling strategy for distribution matching based method.


Recently, SRe$^2$L \citep{yin2023squeeze} proposed to decouple the expensive bi-level optimization and used a three step procedure - first, produce feature mapping; second, generate distilled images; and third generate soft labels. Follow-ups such as \citet{liu2023dataset} and \citet{yin2023dataset_2} brought further improvements on the method. This line of work has achieved impressive performance on large datasets such ImageNet-1K and even ImageNet-21K. 

\textbf{Representing Distilled Dataset}
In contrast to all methods listed so far, a new line of work proposed to distill data into the latent space \citep{deng2022remember,liu2022datset, leehb2022dataset, cazenavette2023generalizing, zhao2022synthesizing}. This line of work proposes to learn the latent code and use decoder(s) to map the latent code back into training images. \cite{zhao2022synthesizing, cazenavette2023generalizing} leveraged pre-trained GANs as the decoder such that only latent code needed to be learned during distillation. \citep{deng2022remember, leehb2022dataset, liu2022datset} trained both latent codes and decoders during the distillation process. 
Our work is mostly similar to IT-GAN \citep{zhao2022synthesizing} in using a generative model to represent distilled data. However, we model the \textit{prototypes themselves} as distributions, allowing for e.g., unlimited sampling from them, and leading to more diverse generation. IT-GAN \citep{zhao2022synthesizing} only showed the feasibility on CIFAR-10 while we scale the idea to TinyImageNet, and ImageNet-1K. Furthermore, we show that by using a distributional framework and a generator trained from scratch, one can achieve a more compact representation of data.

\section{Methodology}

\subsection{Three-Dimensional Evaluation}
\label{sec:metric}
The most important aspect of evaluating data distillation methods is the trade off between the memory footprint (i.e., how large is the distilled dataset) and the recovery accuracy (i.e., can models trained the compressed data achieve comparable performance compared to the original dataset). When it was first proposed by \citet{wang2018dataset}, the distillation task was restricted to finding a set of images $\lbrace \vec{s}_i \rbrace_{i=1}^n$. Same number of prototypes were used for each class along with hard labels. Since storing the corresponding label incurred a trivial cost, IPC (Image Per Class) was sufficient to capture the distilled dataset size in early works. However, two recent trends brought innovations to different ways to store information in the distilled dataset. As a result, the IPC metric no longer reflects the trade-off between storage and recovery accuracy. 


\textbf{Storage Cost} Instead of distilling into pixels, many recent works \citep{deng2022remember, leehb2022dataset, liu2022datset} distill data into a latent space $\mathcal{Z}$ and represent each prototype as a latent code $z \in \mathcal{Z}$. One or multiple decoders are used to map the latent code into original space during downstream training, by trading memory with compute.
On the other hand, images are not the only way one can store information in the distilled dataset. Many methods that scale to ImageNet-1K leverage distilled labels as an additional way to store information. TESLA \citep{yin2023squeeze}, and FRePo \citep{zhou2023dataset} distill prototypes into pixel space and assign one unique soft label to each prototype. Compared to hard labels, storing softmax values incurs a small but non-trivial storage cost. On the other hand, SRe$^2$L and its follow-up work \citep{yin2023squeeze, liu2023dataset, yin2023dataset_2} take a slightly different approach by assigning \textit{multiple} distilled labels to each prototype. For each prototype, different distilled labels correspond to variants of the prototype by applying augmentations. As a result, these work \citep{yin2023squeeze, liu2023dataset, yin2023dataset_2} require the augmentation parameters stored along with the corresponding to distilled labels. 

\textbf{Downstream Training Cost} When training models on the distilled data, using soft labels instead of hard labels, decoding latent codes on-the-fly, and applying augmentations to prototypes all bring additional computation cost during downstream training. Therefore, in addition to storage cost, we also propose to look at the the wall clock time to train models on the distilled data, which we abbreviate as downstream training cost. This training cost can help us gain insights into the memory versus compute trade-off between different distillation methods. However, the primary objective of dataset distillation is to achieve information compression by saving only relevant features needed for a certain task, storage cost should be the primary metric for evaluation and downstream training cost should be a secondary metric. Despite being a secondary metric, downstream task training cost is still relevant because if training models on the distilled data takes too long, the distilled dataset may have limited usability on applications such as continual learning or neural architecture search.


We propose a more comprehensive evaluation process based on the following three metrics:

\begin{enumerate}
[itemsep=2pt,labelindent=0pt,topsep=0pt,parsep=0pt,partopsep=0pt,label=(\roman*), align=left]
\item Total storage cost: being distilled images, prototypes, latent codes, soft labels, augmentations, and/or decoders
\item Downstream training cost: wall clock time it takes to train models on the distilled data 
\item Recovery accuracy: accuracy achieved by model trained on the distilled data 
\end{enumerate}


We perform the comprehensive evaluation on TESLA \citep{cui2022scaling}, SRe$^2$L \citep{yin2023squeeze} and D3 (ours) on ImageNet-1K (resized). TESLA \citep{cui2022scaling} represents SOTA results among existing methods that distill directly into image spaces along with soft labels. SRe$^2$L \citep{yin2023squeeze} represents results among existing methods that distill into images, augmentations and soft labels. Finally, our work (D3) represents results that distill into latent (distributions). Using the new metric, we observe a different landscape that is not captured by IPC, as shown in Figure \ref{fig:pull}. Bi-level optimization-based methods (TESLA and ours) excel at small-scale, extremely efficient dataset distillation while decoupled methods (SRe$^2$L) achieves superior performance at the cost of a larger storage footprint and longer downstream training time. In Appendix \ref{appdx:fig1}, we list further details on the exact storage cost breakdown and a discussion on the storage cost is measured.

\subsection{Distilling into distributions}

Based on the success of distilling into the latent space, we argue that one can further achieve information compression by distilling into a latent distribution. 
To make the problem tractable, we consider a family of distributions $Q_{\theta}$ parameterized by $\theta \in \Theta \subseteq \mathbb{R}^D$. By restricting to the family of parametrizable distributions, we can formulate the problem as an optimization over the finite-dimensional $\Theta$ rather than the infinite-dimensional space of distributions:



We further argue that when we distill into distributions, `distillation' is satisfied if (i) the storage footprint as discussed in the previous section is sufficiently small and (ii) the effective number of samples from $Q$ on which a model needs to be trained is comparable, or lower, than that of training on the original dataset $\mathcal{D}$. In particular, we seek to avoid the two trivial corner-cases $Q_\theta = \tfrac{1}{N} \sum_{i=1}^N \delta_{\vx_i}$, i.e., the uniform empirical measure associated with $\mathcal{D}$, and $Q_\theta \approx P_{\mathcal{D}}$, i.e., learning the full distribution of the original data --- a much harder problem to solve. 

To enforce effective compression, we draw inspirations from Deep Latent Variable models \citep{kingma2019an, kingma2013auto-encoding}, and impose a Gaussian structure to represent the latent distribution. Each Gaussian prior is parameterized by the mean and variance ($\mu_i, \Sigma_i$), and each class can be represented using one or multiple Gaussian priors. Formally, the latent distributions can be expressed as: $Q_{\theta} \sim \mathcal{N} (\mu_i^c, \Sigma_i^c)$, indicating the \textit{i-th} prior for class $c$. During distillation, parameters for the Gaussian distribution $\theta = (\mu_i^c, \Sigma_i^c)$ are learned along with the decoder parameters. We gain fine-grained control over storage cost, downstream training efficiency and distillation quality by varying number of latent priors per class, dimension for the latent Gaussian distribution and the decoder size.



Empirically, we find that the most effective way to scale up the size of distilled dataset and achieve higher recovery accuracy is to first increase the number of latent priors per class. Once we exceed a certain number of priors per class, we also need to use a larger decoder and higher latent dimension to achieve higher distillation quality. See Appendix \ref{appdx:vae_formulation} for a detailed description on latent distribution and Appendix \ref{appdx:decoder} for our decoder architecture. 

\begin{table*}[!t]
\centering
\caption{\textbf{Tiny ImageNet distilled and evaluated on ConvNet} Storage cost is measured in MB and in parenthesis, we annotate the equivalence if storing only images (measured using IPC). In our most compact setting, the storage cost to store the distilled distribution averages to storing less than 1 IPC. N/A: indicates the distillation size is smaller than the minimum size the method can distill.}
\label{tab:tiny}
\resizebox{0.7\columnwidth}{!}{
\begin{tabular}{@{\extracolsep{-4pt}}c cccccccc}
\toprule   
    Storage Cost (MB)  & DM & MTT & LinBa & KFS & FrePo & D3(ours)  \\
     \midrule
     4 \tiny{($\sim$0.5IPC)} & N/A   & N/A   & N/A   & N/A & N/A  & \textbf{\errbar{24.6}{0.2}}\\
     10 \tiny{($\sim$1IPC)} & \errbar{3.9}{0.2}&  \errbar{8.8}{0.3} &  \errbar{16.0}{0.7}   &   \errbar{22.7}{0.2} & \errbar{15.4}{0.3} & \textbf{\errbar{26.0}{0.4}}\\
     100 \tiny{($\sim$10IPC)} & \errbar{12.9}{0.4}&\errbar{23.2}{0.2} &   &    \errbar{27.8}{0.2}  & \errbar{25.4}{0.2} &  \textbf{\errbar{30.5}{0.3}} \\
\bottomrule
\end{tabular}}
\vspace{0.2in}
\centering
\caption{\textbf{Tiny ImageNet distilled and evaluated on different architectures} We use the distribution distilled under the 100MB ($\sim$ 10 IPC) storage cost budget. Our method generalizes well to different architectures. }
\label{tab:tiny_cross}
\resizebox{0.5\columnwidth}{!}{
\begin{tabular}{@{\extracolsep{-4pt}}c cccccc}
\toprule   
    ConvNet (self) & AlexNet & ResNet18 & VGG11 & ViT      \\
     \midrule
     \errbar{30.5}{0.3}  &  \errbar{22.6}{0.5}  &   \errbar{25.7}{0.5}   & \errbar{27.0}{0.2}  & \errbar{15.1}{0.7}\\
\bottomrule
\end{tabular}}
\vskip -0.0in
\end{table*}

\subsection{Federated Distillation}
The challenge to scale data distillation methods to ImageNet-1K comes from the significant memory and computation costs \citep{zhou2023dataset, yin2023squeeze, cui2022scaling}. Our method D3 also suffers from the same challenge. 
To resolve the scaling issue, we propose to use a federated distillation strategy. First, we divide the datasets into $k$ subsets in the class space. Each subset only contains $C/k$ classes, where $C$ denotes the total number of classes in the full set. Then, we perform data distillation independently on each subset. Note that in this step, we train local experts for each sub-task and optimize the distill data on those subtasks, which is simpler than the full classification task. Finally, the distill subsets are aggregated to form the distilled dataset for the full task. For an illustration of our federated distillation strategy, see Figure \ref{fig:pull_demo}. 

Since each subset has only been trained by local experts for each subtask (i.e, classify only $C/k$ classes as opposed to all $C$ classes), one certainly would expect the federated strategy to yield sub-optimal results compared to directly distilling on the full dataset. In section \ref{sec:federated} we confirm such intuition. However, we observe that dataset distilled on those simpler subtasks transfers relatively well to the full task. This nice generalization property allows us to distill ImageNet-1K in a highly parallelized fashion while achieving SOTA results. 



\subsection{Training Objective}
Building on the foundations of existing data distillation techniques, we introduce a learning objective compromised of two distinct terms. The first term is derived from Matching Training Trajectories (MTT) proposed by \citet{cazenavette2022dataset}. The second term in our objective aims to minimize the Maximum Mean Discrepancy (MMD) between the true dataset and our learned dataset distribution. Different from the formulation used in DM \citep{zhao2023dataset}, we use a set of Reproducing Hilbert Kernels (RHKS) for the MMD computation to fully leverage the power of MMD. We first map the pixel space to latent feature space using trained experts. For model simplicity and training efficiency purposes, we recycle the experts used in MTT to generate feature mappings. We then use a mixture of Radial Basis Function (RBF) kernels $k(x, x') = \sum_{q=1}^K k_{\sigma_q}(x, x')$, where $k_{\sigma_q}$ represents a Gaussian kernel with bandwidth $\sigma_q$. We choose a mixture of $K=5$ kernels with bandwidths $\lbrace 1, 2, 4, 8, 16\rbrace$. The hyperparameter choice is inspired by MMD GANs \citep{binkowski2018demystifying, li2017mmd}. See Appendix \ref{appdx:loss_term} for a full description of the training objective.

\section{Experiments}
In pursuit of impartial comparisons with existing data distillation methodologies, we align all our design choices with existing work. We use ConvNet for data distillation on all datasets. We evaluate the recovery accuracy on five randomly initialized neural networks and report mean and standard deviation. We provide a detailed description of datasets, experiment setup and hyper-parameters in Appendix \ref{appdx:experiment_details}. We compare our methods on competitive baselines that distill into pixel space, including MTT \citep{cazenavette2022dataset}, TESLA \citep{cui2022scaling} , concurrent work DataDAM \citep{sajedi2023datadam}, FRePo \citep{zhou2023dataset}, FTD \citep{du2022minimizing}, and DM \citep{zhao2023dataset}. We also compare our method on competitive baselines that distill into latent space, including LinBa \citep{deng2022remember} and KFS \citep{leehb2022dataset}.

\subsection{Quantitative Results}
\label{sec:quant_results}

\textbf{Benchmark Performances} We apply the federated distillation strategy on \textbf{ImageNet-1K} by breaking down the dataset into 2 and 5 sub-tasks. To scale up the distilled distributions, we use 1, 2 and 10 latent priors per class and scale up decoder sizes accordingly. We report results on both ConvNet4D and ResNet18 (cross-architecture generalization) with our three-dimensional evaluation metric in Figure \ref{fig:pull} and a tabular version can be found in Appendix \ref{appdx:fig1}. Our method outperforms TESLA \citep{cui2022scaling}, DataDAM \citep{sajedi2023datadam} and FRePo \citep{zhou2023dataset} on ConvNet under 100MB and 500MB storage budget. Furthermore, our method can also distill a distribution under 25MB (0.5IPC) storage budget, which outperforms existing work under 50MB (1PC) storage budget. 
Our method distills a more compact dataset through distributional representation, however, when we evaluate on the downstream task training cost, we can see that that our compact representation comes at a (small) compute cost. On average, models trained on our distilled distribution takes more training iterations to converge compared to fixed-output methods and generating images on-the-fly costs additional small but non-negligible compute time. We report \textbf{TinyImageNet} results in Table \ref{tab:tiny}. We use 2, 5, and 10 latent priors for three storage costs and a larger decoder for the last one. Similar to ImageNet-1K, our method can distill distribution that achieves SOTA performance with small storage costs. Additionally, we report CIFAR-100, CIFAR-10 and the two ImageNet subsets in Appendix \ref{appdx:add_results}. In Appendix \ref{appdx:decoder_list}, we include a full list of decoder hyper-parameters and exact storage cost for all the experiments listed above. 
\begin{wrapfigure}[20]{r}{0.4\textwidth}
\centering
\includegraphics[width=0.4\columnwidth]{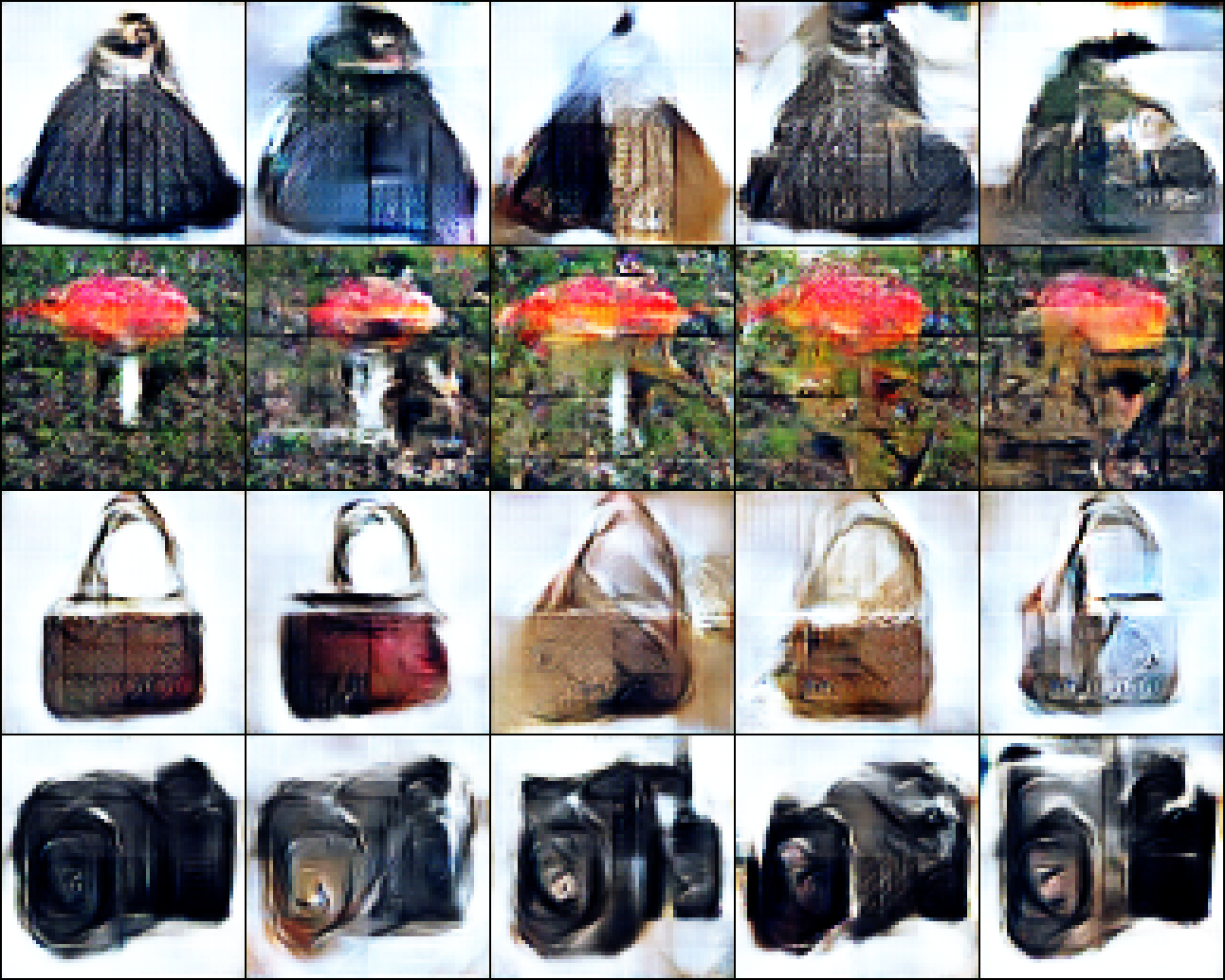}
\caption{\textbf{Visualization of distilled mean and variations for four classes from ImageNet-1K} \textit{ first column:} typical images for each class generated by passing the mean to the decoder to generate the sample. \textit{second column onwards:} variations by sampling from the corresponding latent distribution}
\label{fig:variations}
\end{wrapfigure}

\textbf{Cross-Architecture Generalizability} Figure \ref{fig:pull} left panel, and Table \ref{tab:tiny_cross} shows that our distill distributions, trained on ConvNet, generalizes well to ResNet18 and other architectures. We report additional cross-architecture generalization results for CIFAR-10, ImageNette and ImageWoof, and compare our method with existing work in Appendix \ref{appdx:add_results}. Our method does not generalzie well to Vision Transformers - it is a common yet not resolved pitfall faced by many data distillation algorithms \citep{cazenavette2023generalizing, liu2023dataset}.

\subsection{Latent Space Analysis}
Recall that we impose a Gaussian structure to the latent distribution space, and the motivation behind distillation is to save only relevant information from a dataset for a specific task (image classification, in our case). Through the distilled distribution, we can visualize the Gaussian space to reach some qualitative understanding on the `salient' features that are essential for image classification task. In Figure \ref{fig:variations}, we visualize the prototype distribution of distilled ImageNet-1K under 500MB ($\sim$ 10 IPC) storage budget. Specifically, we visualize the average (first column) of four randomly chosen classes and their corresponding variations (second column onwards). Qualitatively, we observe that the typical (i.e., mean) sample is more interpretable and higher quality compared to its variations. 
\begin{figure}[!t]
\vspace{-0.1in}
\centering
\centerline{
\includegraphics[width=0.4\columnwidth]{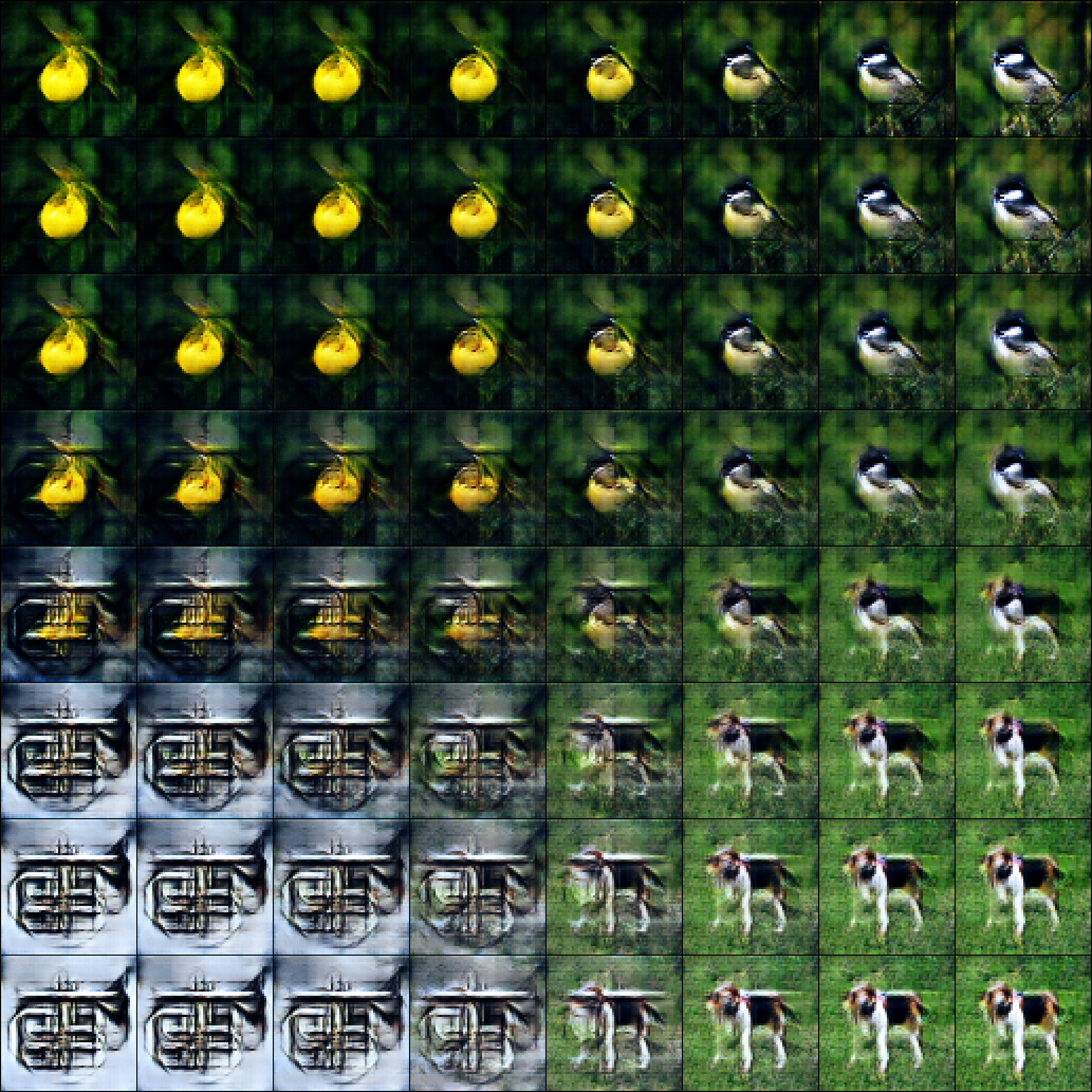}
}
\caption{\textbf{Visualization of the latent Gaussian space by interpolating priors for four classes from ImageNet-1K} The four corners are the mean for each class. We linearly sample the Gaussian space between classes and pass into the decoder to generate interpolated images.
}
\label{fig:latent_space}
\vspace{-0.2in}
\end{figure}
We also visualize four randomly chosen classes (four corners) and the inter-class ``distributions" by linearly interpolate the Gaussian space between (the rest), shown in Figure \ref{fig:latent_space}. 

\subsection{Federated Distillation}
\label{sec:federated}
\textbf{Ability to task-generalize} When we perform federated distillation, we are essentially distilling datasets on simpler subtasks (i.e., classification on fewer classes). To understand the extent to which breaking down distillation tasks can impact the overall performance on the distillation process, we conduct the following experiments on TinyImageNet. 

First, we perform federated data distillation by dividing the dataset into two subsets, first one containing the first hundred classes and the second one containing the rest (second hundred classes). The full distilled dataset distribution is obtained from aggregating the two distributions. To form a comparison, we perform dataset distillation directly on the full dataset using the same-sized decoder, and the same settings for the latent prior distributions (i.e., same dimension for the latent Gaussian distribution, and same number of latent priors per class). 

Experiment results are reported in Table \ref{tab:task_transfer}. Rows indicates the dataset that distillations are performed on, and the column indicates the dataset that distilled distributions are evaluates on. First row summarizes the federated distillation outcome: each subtask achieves $\sim 27\%$ recovery accuracy. We then evaluate the aggregated distributions on the full task, which achieves $21.9\%$ recovery accuracy (row 1, col 2). In comparison, the second row summarizes results for the non-federated counterpart: directly distilling on the full set achieved higher accuracy $24.6\%$ than the federated outcome. In this experiment, we observe that the federated distillation only slightly underperforms the non-federated version. In addition, we notice that the data distribution distilled on the full dataset outperforms the federated counterpart (row 2, col 1) when evaluated on the subtask. 

\begin{wraptable}[10]{r}{0.55\textwidth}
\centering
\caption{\small{\textbf{Federated distillation compared to full task distillation on TinyImageNet} row indicates the dataset being distilled on and column indicates the dataset being evaluated on.}}
\label{tab:subtask_performance}
\begin{tabular}{@{\extracolsep{-2pt}}ccccc}
\toprule   
      & (Subset 1, Subset 2) & Full \\
     \midrule
       (Subset 1, Subset 2) & (\errbar{27.6}{0.6}, \errbar{27.2}{0.5})& \errbar{21.9}{0.6}   \\
       Full  & (\errbar{32.2}{0.3}, \errbar{33.4}{0.4}) & \errbar{24.6}{0.3}  \\
\bottomrule
\end{tabular}
\vskip -0.5in
\end{wraptable}


\textbf{Impact factors} 
To understand whether the ability of distilled data to task generalize is sensitive to the distillation training objective and/or the use of distributional outcome, we further experiment on TinyImageNet using the same set-up as above. In this set of experiments, we perform data distillation using different training objectives, using distributional or fixed outputs to distill each of the TinyImageNet subsets, shown in Table \ref{tab:task_transfer}. 

In Section \ref{sec:ablation_loss}, we perform a more detailed ablation study on the impact of training objectives and distributional representation on distillation outcomes. Here, we are only interested in examining whether and to what extent those factors impact the ability for the distilled data to task generalize. For fixed outputs, we repurpose latent distributional priors as latent codes and simply distilling only the mean. From Table \ref{tab:task_transfer}, we observe that the relative transfer performance is not sensitive to different training objectives. On the other hand, when we restrict ourselves to fixed outputs (i.e., without distributional representation), the relative task transfer ability suffered by a non-trivial amount. The performance drop is most evident when we allow fewer fixed latent codes per class. However, as we increased the number of fixed images, the task transfer ability converge to the distributional version. This observation indicates that our federated distillation scheme could potentially be generalized to other data distillation methods. 
\begin{table}[ht]
\centering
\caption{\textbf{Federated distillation on TinyImageNet using different training objectives, and using distributional or fixed representation} \textit{Subset 1, Subet 2} is recovery accuracy on the subtask and \textit{full} is recovery accuracy by aggregating the two distill datasets and evaluating on the full set. Chg$\%$ computed by \texttt{full}/\texttt{avg(subset 1, subset 2)}.}
\label{tab:task_transfer}
\vspace{0.1in}
\resizebox{0.7\columnwidth}{!}{
\begin{tabular}{@{\extracolsep{-2pt}}cccccc}
\toprule   
    Loss Term &Distributional &Subset 1($\%$) & Subset 2($\%$) & Full($\%$)  & Chg$\%$       \\
     \midrule
     MTT   & Yes & \errbar{12.7}{0.2} & \errbar{18.4}{0.4} &  \errbar{13.4}{0.4}&86$\%$\\
     MMD &Yes & \errbar{25.4}{0.4} & \errbar{26.1}{0.5}  & \errbar{20.4}{0.1} &80$\%$\\
     Both & Yes & \errbar{27.6}{0.3} & \errbar{27.2}{0.5}  & \errbar{21.9}{0.6} &80$\%$\\
     Both & No (5 IPC) & \errbar{10.28}{0.5} & \errbar{12.38}{0.2} & \errbar{7.5}{0.5} & 66$\%$\\
     Both & No (10 IPC) & \errbar{19.58}{0.4} & \errbar{17.68}{0.5} & \errbar{13.9}{0.2} & 74$\%$\\
     Both & No (20 IPC) & \errbar{27.5}{0.3} & \errbar{25.32}{0.4} & \errbar{20.5}{0.6} & 78$\%$\\
\bottomrule
\end{tabular}}
\end{table}

\begin{table}[ht]
\centering
\vspace{0.1in}
\caption{\textbf{Performance of federated distillation with different subtask sizes on ImageNet-1K.} Increasing the number of subtasks has a negative impact on the performance of federated distillation}
\label{tab:task_breakdown}
\vspace{0.1in}
\resizebox{0.5\columnwidth}{!}{
\begin{tabular}{@{\extracolsep{-2pt}}cccc}
\toprule   
    Subset size & $\#$ Tasks& Decoder Size &  Accuracy       \\
     \midrule
     100  & 10 & S & \errbar{9.7}{0.2} \\
     100  & 10 & M & \errbar{9.6}{0.3}\\
     200  & 5 & S & \errbar{10.6}{0.3} \\\
     200  & 5  & M & \errbar{13.8}{0.5}\\
     500  & 2 & L  &\errbar{14.7}{0.5} \\
\bottomrule
    \end{tabular}}
\end{table}


\textbf{Number of Subtasks} To understand to what extent the number of subtasks negatively impacts the federated distillation strategy, we experiment three division sizes on ImageNet-1K: 10 sub-tasks (100 classes for each task), 5 sub-tasks (200 classes for each task) and 2 sub-tasks (500 classes for each task). We use two different decoder sizes for the first two and a large one for the later \footnote{smaller and medium sized decoder failed to converge on 500-class subtask}. For the same decoder size, we keep the hyper-parameters same for the latent distribution (same latent dimension, and same number of priors per class). As we divide into smaller sub-tasks, the distillation process becomes more parallelizable and distilling on each subtask becomes simpler. Not surprisingly, it comes at a cost of worse generalization capability for the distill data, which leads to worse recovery accuracy for the full task. Table \ref{tab:task_breakdown} shows that increasing the number of tasks has a negative impact on the overall performance.

\section{Ablation Study}
\subsection{Distributional Outcome}
\label{ablation:distribution}
\begin{figure}[!th]
\vspace{-0.1in}
\begin{center}
\centerline{
\includegraphics[width=0.7\textwidth]{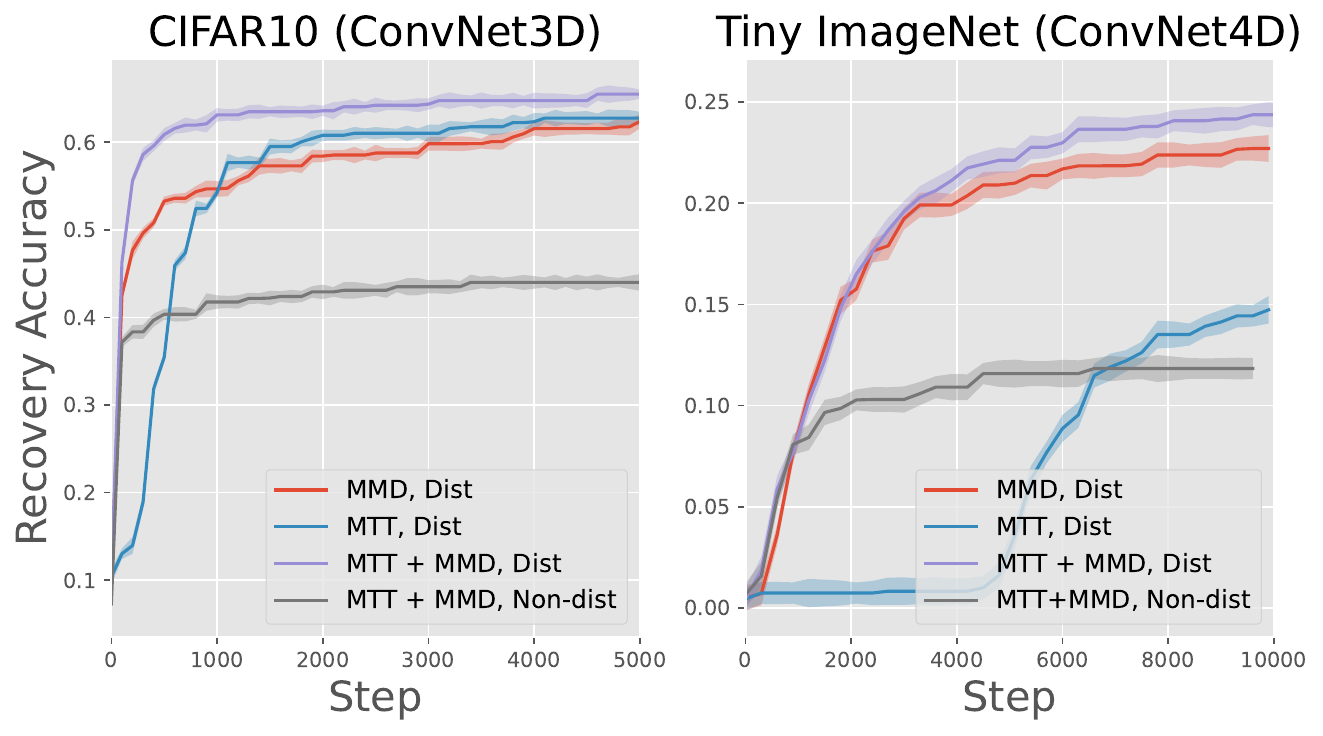}
}
\caption{\textbf{Ablation study on dual training objective and distributional representation} Both distributional representation and dual training objective are essential for the performance of our method.}
\label{fig:loss_term_ablation}
\end{center}
\vspace{-0.2in}
\end{figure}

\textbf{On training accuracy} Using the same training objective and using the same decoder setting, we experiment disabling the distributional outcome. By allowing a distributional representation, a more diverse set of samples are generated on-the-fly. As a result, the distilled distribution reaches a higher distillation quality compared to its non-distributional counterpart. We experiment on CIFAR-10 and TinyImageNet, shown in Figure \ref{fig:loss_term_ablation}. \textbf{On prototype quality} We also visualize samples from distilled TinyImageNet outcomes above (with and without distributional representation) in Figure \ref{fig:distributional_ablation}. We observe that the distributional objective makes the distilled data more interpretable. 

\subsection{Loss Term Contribution}
\label{sec:ablation_loss} 
\begin{figure}[!t]
\begin{center}
\centerline{\includegraphics[width=0.4\columnwidth]{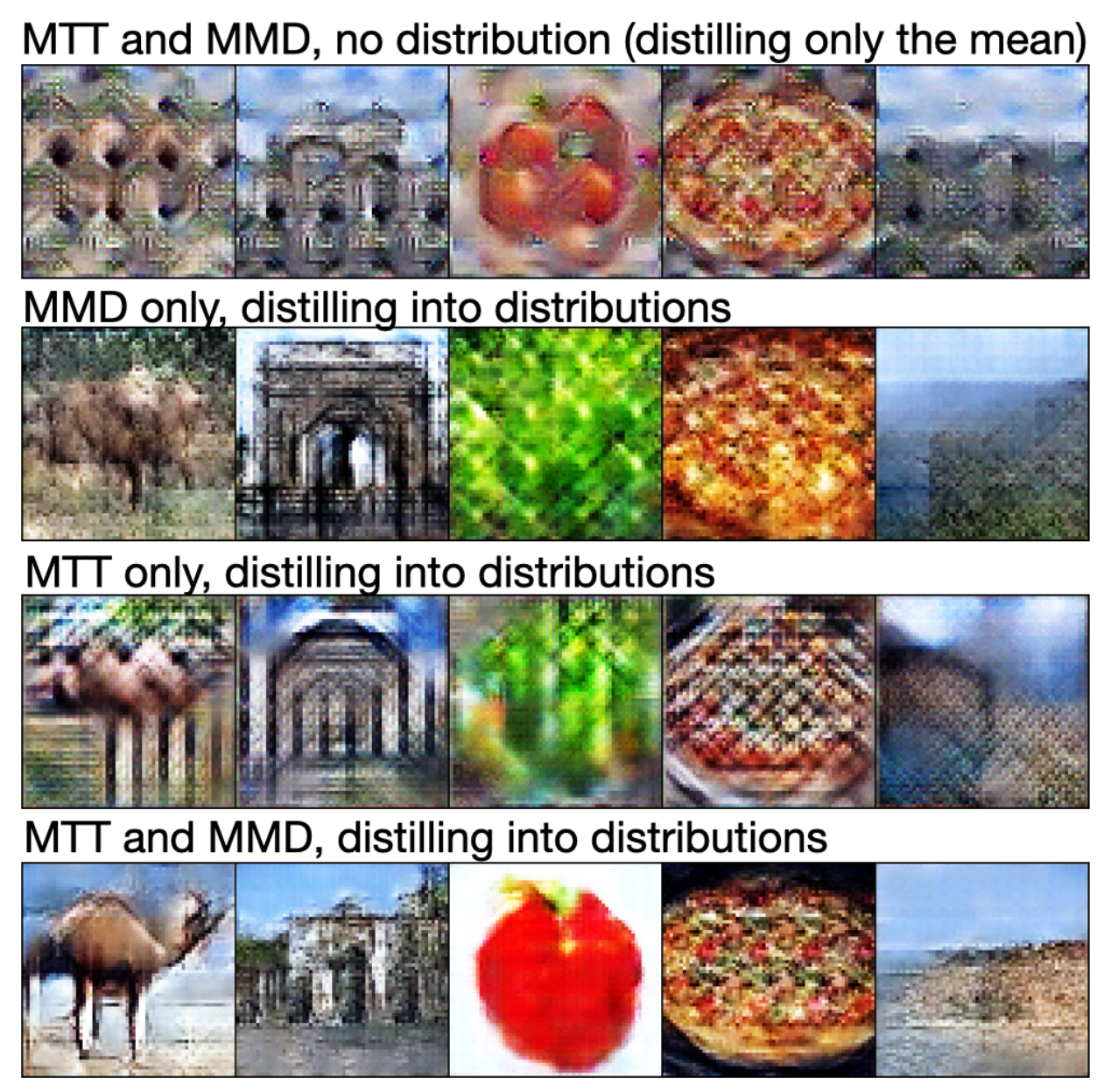}}
\caption{\textbf{Visualization of distilled samples from five classes using different training objectives and fixed or distributional representation on TinyImageNet} for distributional outcomes we visualize the mean.}
\label{fig:distributional_ablation}
\end{center}
\vspace{-0.2in}
\end{figure}

To study the effect of combing two objectives, we perform distillation on CIFAR-10 and TinyImageNet with either loss terms while keeping the decoder hyper-parameters constant, results shown in Figure \ref{fig:loss_term_ablation}. The dual training objective yields superior performance than using either one stand alone. However, using the MMD or MTT objective alone could already achieve good results, depending on the dataset. While performing distillation, we observe that the dual objective consistently outperform using one alone. From visualizations in Figure \ref{fig:distributional_ablation}, the combined objective yields more interpretable results than using either alone. While it might be possible that one can further simplify the training objective by only using one of them, we keep the dual objective based on the above observations. 

\subsection{Distilled Labels}
\begin{wrapfigure}[17]{r}{0.5\textwidth}
\vspace{-0.2in}
\begin{center}
\centerline{
\includegraphics[width=0.5\columnwidth]{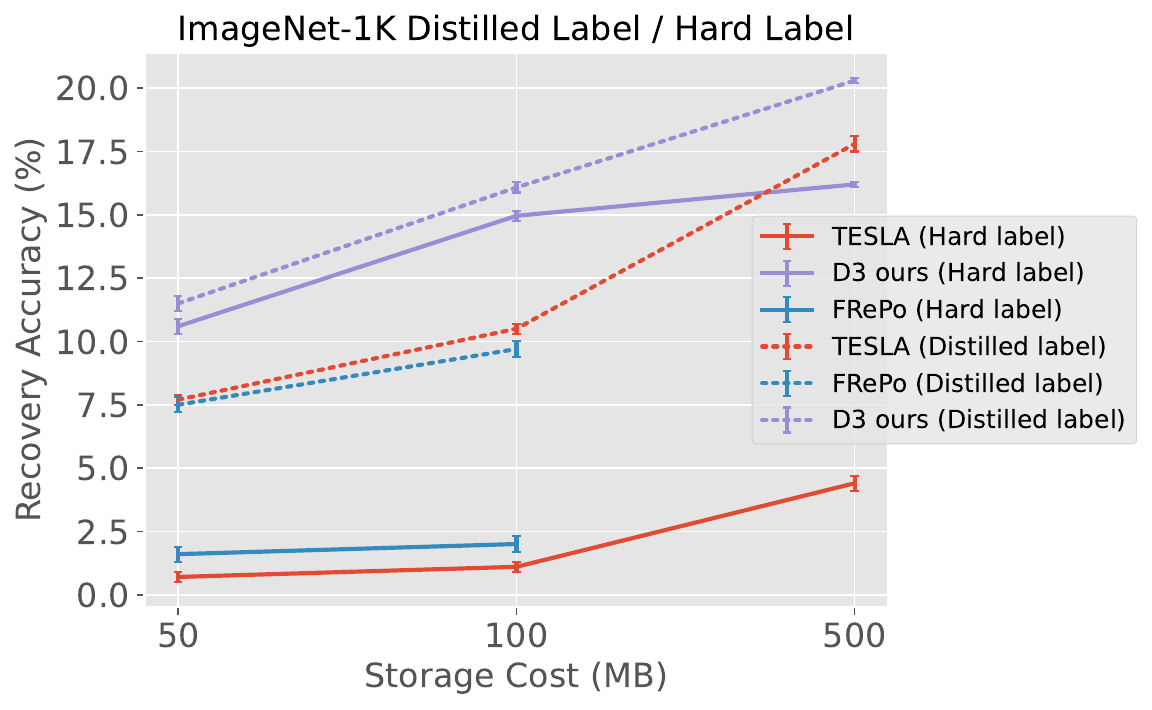}
}
\caption{\textbf{ImageNet-1K recovery accuracy using distilled labels or hard labels} Our method still performs well even without using distilled labels.}
\label{fig:nosoft_compare}
\end{center}
\end{wrapfigure}

Similar to prior works, we also find that the use of distilled labels brings additional benefit to the dataset distillation. We use softmax values generated by pretrained experts as distilled labels, an intuitive strategy already used by \cite{cui2022scaling, zhou2023dataset}. However, since we distill into distributions, we use the mean of every latent prior to generate soft labels. In practice, we find that the distilled labels for each mean work well even for randomly generated samples. Different from existing work, our distilled distribution is more robust against having only hard labels - our method significantly outperforms TESLA \cite{cui2022scaling} and FRePo \cite{zhou2023dataset} when only hard labels are used, shown in Figure \ref{fig:nosoft_compare}. 

\section{Conclusion}
In this paper, we first made a key observation that existing work distilling into explicit prototypes \textit{and} distilled labels incurred unexpected storage cost and post-distillation training time, both of which could not be captured the conventional metric used dataset distillation. We proposed to evaluate distillation methods on a three-dimensional metric that captures the total storage cost and the test-time runtime efficiency. 

We also proposed \textit{Distributional Dataset Distillation}, which encodes the data using minimal sufficient per-class statistics and a decoder, resulting in a compact representation that is more memory-efficient representation of distilled data compared to prototype-based methods. To scale up the process of learning these representations, we proposed a federated distillation strategy, which decomposes the dataset into subsets, distills them in parallel using sub-task experts and then re-aggregates them. We evaluated our methods against other SOTA methods and showed that our method achieved state-of-the-art results on TinyImageNet and ImageNet-1K. For future work, we aim to scale our results to larger datasets and to higher distillation quality.


\vskip 0.2in
\bibliography{paperpile}
\bibliographystyle{iclr2024_conference}

\appendix
\clearpage
\section{Methodology (Extended)}
\subsection{Further details on Figure \ref{fig:pull}}
\label{appdx:fig1}
In this section, we provide details for Figure \ref{fig:pull} in Tables \ref{tab:imagenet} and \ref{tab:storage_expanded}. First, in Table \ref{tab:imagenet} we compare our method with SOTA distillation methods at various storage costs. All methods perform distillation on a ConvNet architecture. Additionally, we evaluate our method and TESLA on ResNet18 to examine cross-architecture generalization. We also annotate storage cost with the equivalence of storing images as distilled dataset. Table \ref{tab:storage_expanded} lists all the methods we used to generate Figure \ref{fig:pull}. We report IPC whenever applicable.




\begin{table*}[h]
\centering
\caption{\textbf{ImageNet-1K Performance comparison for SOTA methods aligned on storage budget} Storage cost is rounded (See Table \ref{tab:storage_expanded} for exact storage breakdown). N/A: indicates the smallest size the method can distill into is larger than the corresponding size.}
\label{tab:imagenet}
  \resizebox{0.85\columnwidth}{!}{

\begin{tabular}{@{\extracolsep{0pt}}c ccccccc}
\toprule       
    &  \multicolumn{4}{c}{ConvNet } &  \multicolumn{3}{c}{ResNet18}    \\
     \cmidrule(lr){2-5} \cmidrule(lr){6-8} 
        Storage Cost (MB)  & FrePo & DataDAM & TESLA & Ours & Random  & TESLA  & Ours      \\
     \midrule
     25  \tiny{($\sim$0.5IPC)} & N/A & N/A& N/A & \textbf{\errbar{11.5}{0.5}} & N/A  & N/A  &  \textbf{\errbar{9.7}{0.8}} \\
     50  \tiny{($\sim$1IPC)}& \errbar{7.5}{0.3} & \errbar{2.0}{0.1}& \errbar{7.7}{0.2}  &  & \errbar{0.5}{0.1}  &  \errbar{6.2}{0.5}   &  \\
     100  \tiny{($\sim$2IPC)} &  \errbar{9.7}{0.2} & \errbar{2.2}{0.1}& \errbar{10.5}{0.2}  & \textbf{\errbar{17.4}{0.7}} & \errbar{0.6}{0.1} &  \errbar{9.1}{1.5}   &  \textbf{\errbar{16.0}{0.7}} \\
     500 \tiny{($\sim$10IPC)} & &  \errbar{6.3}{0.1} &\errbar{17.8}{1.3} & \textbf{\errbar{20.3}{0.9}} &\errbar{3.6}{0.1} & \errbar{15.3}{1.3}   & \textbf{\errbar{18.2}{0.6}}  \\ 
     3000 \tiny{($\sim$50IPC)} & & \errbar{15.5}{0.2}&\errbar{27.9}{1.2} & & \errbar{15.3}{2.3} & \errbar{23.2}{0.9} &  \\

\bottomrule
\end{tabular}}
\end{table*}

\begin{table*}[h]
\centering
\caption{\textbf{Details on storage cost breakdown, downstream task training cost and recovery accuracy distilled on ImageNet-1K and evaluated on ResNet18} Input Storage Cost refers to distilled synthetic images for prototype-based methods, and refers to latent prior and decoder for D3 (ours). DTC stands for Downstream task training cost defined in Section \ref{sec:metric}}
\label{tab:storage_expanded}
  \resizebox{\columnwidth}{!}{
\begin{tabular}{@{\extracolsep{-2pt}}c ccccccccc}
\toprule   
    Method & Distill Arch & IPC & Input Storage  (MB)   &  Label Storage  (MB)   &  Accuracy  (\%)&  DTC (min) \\
     \midrule
     
      TESLA & ConvNet & 1 & 58 &  4 & \errbar{6.2}{0.5}& 8 \\ 
      TESLA & ConvNet & 2 & 116 &  8 & \errbar{9.1}{1.5} &  17 \\ 
      TESLA & ConvNet & 10 & 579 &  38 & \errbar{15.3}{0.8} &  60\\ 
      TESLA & ConvNet & 50 & 2897 &  238 & \errbar{23.22}{0.9}  & - \\ 
      
      SRe2L (orig) &ResNet18 &  1 &  696 & 1229  & \errbar{2.9}{0.2}&  50\\ 
      SRe2L (orig) &ResNet18 &  10 & 6948   & 6145  &2\errbar{1.3}{0.6} & 250 \\ 
      SRe2L (orig) &ResNet18 &  50 &  34764  & 30725  & \errbar{46.8}{0.2}& 600 \\ 

      SRe2L (resize) &ResNet18 &  2 &  116 & 1229  & \errbar{1.2}{0.1}&  25 \\ 
      SRe2L (resize)&ResNet18 &  10 & 579   & 6145  &\errbar{10.7}{0.5} & 40 \\ 
      SRe2L (resize) &ResNet18 &  50 &  2897  & 30725  & \errbar{29.0}{0.5}& 175 \\ 
      
      D3(Ours) & ConvNet & N/A & 17 & 4  & \errbar{8.9}{0.8} & 20 \\ 
      D3(Ours) &ConvNet &  N/A & 76 & 8  & \errbar{15.5}{0.7} & 40 \\ 
      D3(Ours) & ConvNet & N/A & 440 &  38 & \errbar{17.5}{0.6}& 60 \\ 

      
\bottomrule
\end{tabular}}
\end{table*}
\textbf{TESLA} \cite{cui2022scaling}: We replicated results for 1/2/10 IPC settings to produce the results above. For the 10 IPC setting, \citet{cui2022scaling} reported much lower performance on ResNet18 ($7.7\%$), and we used our reproduced results with higher accuracy($15.3\%$). We failed to repliate results for 50 IPC and obtained results directly from authors and therefore do not have downstream training cost estimate. 
    
\textbf{SRe$^2$L} \cite{yin2023squeeze}: SRe$^2$L was originally implemented on higher resolution ImageNet-1K ($224\times 224$). We replicated SRe$^2$L using hyperparameters provided by authors, which set of results are denoted as ``original" above. We also re-implemented the pipeline on resized ImageNet-1K ($64\times 64$), which set of results are denoted as ``resized" above. In the original implementation, the image was saved in \texttt{.jpeg} format and the soft label and the augmentation parameter was saved in \texttt{fp16} format. We recomputed the storage cost assuming both images, augmentation parameters and soft labels were saved as floating point tensor format. In the default hyperparameter setting, SRe$^2$L generated 300 distilled labels for each prototype by applying augmentations to each prototype. As a result, both the labels and augmentation parameters needed to be saved, causing a rather large storage cost on label storage. \footnote{50 IPC distilled dataset \citet{yin2023squeeze}: \small{\href{https://huggingface.co/datasets/zeyuanyin/SRe2L/tree/main}{\texttt{https://huggingface.co/datasets/zeyuanyin/SRe2L}}}}
   
    
    


\textbf{Computing storage cost} To facilitate a direct and meaningful comparison between methods that use different distillation approaches, we quantify the total storage cost of each method, including all generated artifacts that are needed to reproduce the distilled dataset (decoder weights, images prototypes, soft labels, augmentations parameters etc). For prototypes and images, we measure their memory footprint when saved in single-precision floating-point tensor format (\texttt{fp32}). For our method, which distills into latent priors and decoder(s), we save all decoder and latent prior parameters again in \texttt{fp32} format (saving the entire $\texttt{state\_dict}$ for the decoder). Likewise, we measure memory footprint for distilled soft labels and augmentation parameters in the same \texttt{fp32} format. We report all computed storage cost in Megabytes (MB) rounded to closest integer value. While one could potentially achieve better image compression rates saving images into alternative formats (\texttt{.jpeg} for example), this would prevent an apples-to-apples comparison. Moreover, similar improvements could be achieved for distilled model parameters via quantization or     compressed (e.g., \texttt{.zip}) storage. However, all these additional steps (if used) bring further complications to the discussion without bringing useful insight into using storage cost as a metric for dataset distillation. As a result, we decide the most natural and fair comparison is to assume floating point tensor format as the unified way of storing distilled data. 

\textbf{Downstream training cost} All test-time runtime experiments are run using two NVIDIA A100-SXM4-40GB GPUs with data parallelism to ensure fair comparison. We use default parameters or hyperparameters provided whenever available. For TESLA, we used learned learning rate for Convnet, default learning rate for ResNet18, default learning rate scheduler, default training epochs (1000 epoch) and default batch size (i.e., batch size = IPC). For SRe$^2$L we used default learning rate, default training epochs (300 epoch) and default batch size. For D3 (our method), we used learned learning rate for Convnet, default learning rate for ResNet18, default batch size (i.e., batch size = latent prior per class) and default training epochs (2000 training epochs). We allowed early termination if all above methods converged earlier than the default epoch setting. However, for all methods, using default parameters only provides a rough estimate for the downstream training cost, and it may be possible to further optimize for downstream training cost with hyperparameter tuning. While it should only be used as a secondary metric to evaluate data distillation methods, we hope future work could provide more details on this metric when reporting results.

\subsection{Parametrizing the Distilled Distribution}
\label{appdx:vae_formulation}
Our parametrization of the distilled distribution takes inspiration from Deep Latent Variable Models (DLVMs, \citet{kingma2019an}), a flexible family of statistical models that combine the foundational principles of variational inference with the approximation power of deep neural networks. DLVMs have the advantage of encoding information into a latent (typically lower-dimensional) space and provide a method to map from it to data space, enabling flexible generation. Here, we leverage the benefits of latent variable models but adapt them to the distributional formulation of the problem introduced in the previous section. Concretely, we represent the distilled dataset directly as a probability distribution, parametrized as a DLVMs. We model the distilled distribution in a variational form as 
\begin{align*}
    Q_{\mathcal{S}}^{\theta}(\vx) = \int Q_{\mathcal{S}}^{\theta} (\vx\vert \vz) p(\vz)\diff \vz,
\end{align*}
where $Q_{\mathcal{S}}^{\theta}(\vx\vert \vz)$ is parametrized by a neural network and maps the prior $z$ to the distilled samples $\vx$.

We may assume that each class shares one prior distribution, $p(\vz \vert c) \sim \gauss{\mu_c}{\Sigma_c}$, with the parameter pair $(\mu_c, \Sigma_c)$ learned during the data distillation process. Therefore, data points for each class follow a multivariate Gaussian distribution. To scale to larger distillation sizes, we use multi-modal distribution by allowing multiple Gaussian priors for each class. We refer to the number of Gaussian priors as latent priors per class. Under the multi-modal set up, for each class $c$, the set of all possible prior distributions $\lbrace \gauss{\mu^{i}_c}{\Sigma^{i}_c}\rbrace _{i=1}^{\mathrm{PPC}}$ follows a uniform distribution. 

\subsection{Training Objective}
\label{appdx:loss_term}
\paragraph{MTT Loss}
Expert trajectories are training trajectories generated from training neural networks on the full training set. At each distillation step, we initialize a student network that has the same architecture as the experts. The student network's initialization weight $\vw^{Q}$ is sampled from the experts training trajectory by randomly selecting an expert and a random iteration $t$, such that $\vw^{Q}_{t} = \vw^{\mathcal{D}}_{t}$. We perform $N$ gradient updates on the student network using data drawn from the distilled distribution: 
\begin{align*}
    \mathrm{for}\, n &= 0....N-1:  \vw^{Q}_{t+n+1} = \vw^{Q}_{t+n} - \alpha \nabla \loss(Q; \vw^{Q}_{t+n}), Q\sim Q_{\mathcal{S}}^{\theta} 
\end{align*}

We then collect expert parameters from $M$ training updates after iteration $t$, which denote as $\vw^{\mathcal{D}}_{t+M}$. The distance between the updated student parameters and the updated expert parameters is quantified using normalized squared error: 
\begin{align*}
    D_\texttt{MTT} = \frac{\norm{\vw^{Q}_{t+N} - \vw^{\mathcal{D}}_{t+M}}^2_2}{\norm{\vw^{\mathcal{D}}_{t} - \vw^{\mathcal{D}}_{t+M}}^2_2}
\end{align*}

\paragraph{MMD Loss} 
We use a set of Reproducing Hilbert Kernels (RHKS) for the MMD computation to fully leverage the power of MMD. Since we only have access to the distilled distribution $Q_{\mathcal{S}}^\theta$ but not the training data distribution $P$, we use the empirical MMD measure:
In  general, given random variable $X = \lbrace x_1, ..., x_n\rbrace \sim \mathbb{P}$ and  $Y = \lbrace y_1, ..., y_m \rbrace \sim \mathbb{Q}$, the unbiased estimator of the MMD measure is \cite{li2017mmd}:
\begin{align}
\label{eqn:mmd_data}
    \mathrm{\widehat{MMD}}^2(X, Y) = \frac{1}{\binom{n}{2}}\sum_{i\neq j}^n k(x_i, x_j)  -\frac{1}{m n}\sum_{i=1}^n\sum_{j=1}^m (x_i, y_j) + \frac{1}{\binom{m}{2}}\sum_{i\neq j}^m k(y_i, y_j) 
\end{align}
We also map the pixel space to latent feature space. 
For model simplicity and training efficiency purposes, we recycle the experts used in MTT to generate feature mappings, and denote them as $\psi(\cdot)$. 
Inspired by MMD GANs (see \cite{li2017mmd, binkowski2018demystifying}), we use a mixture of Radial Basis Function (RBF) kernels $k(x, x') = \sum_{q=1}^K k_{\sigma_q}(x, x')$, where $k_{\sigma_q}$ represents a Gaussian kernel with bandwidth $\sigma_q$. We choose a mixture of $K=5$ kernels with bandwidths $\lbrace 1, 2, 4, 8, 16\rbrace$.

To encourage distribution matching with the original dataset, we penalize large MMD: 
\begin{align}
\label{eqn:mmd_def}
    \mathcal{L}_{\texttt{MMD}} = \sum_{c=1}^{C}\mathrm{\widehat{MMD}}^2(\psi(\mathcal{D}_c), & \psi(\mathcal{S}_c)),
\end{align}
where the $\mathrm{\widehat{MMD}}^2$ computation is defined in Eqn. \ref{eqn:mmd_data}. $\mathcal{D}_c$ and $\mathcal{S}_c$ simply refers to the subset of the training data or distilled data with class label $c$.

\section{Details on the decoder}
\label{appdx:decoder}
Our decoder is adopted from the decoder part of the VAE designed by \cite{kingma2019an}, with small modifications. First, we project the latent $z$ in to a $k$ dimension feature vector, which is then fed into a sequence of 2D \texttt{ConvTranspose} blocks. Each of the decoder block contains a \texttt{ConvTranspose} layer followed by a \texttt{BatchNorm} layer and a \texttt{LeakyReLU} activation. For larger decoder, we increase the latent dimension, and consequently the size of \texttt{ConvTranspose} blocks. After the those blocks, there is a 2D convolutional layer followed by a \texttt{tanh} activation. The exact dimension of the convolution layer differs by image output size. The original VAE was designed only for images with size $32\times32$, and used only 3 blocks. We also increase the number of deconv blocks for larger datasets. 

\begin{table*}[h]
\centering
\caption{\textbf{Architecture and hyperparameter details for the three sizes of decoders we used} \textit{Total parameters} are counted in millions. $\#$\textit{Blocks} indicates the number of convolutional blocks.}
\label{tab:decoder_archi}
  \resizebox{0.7\columnwidth}{!}{
\begin{tabular}{@{\extracolsep{-4pt}}c ccccccc}
\toprule   
    Size  & $\#$ Blocks & Latent Dimension  &  Total Params(M)  & Output Image Size \\
     \midrule
     S & 5 & 64  & 0.75 &$32\times32\times3$  \\ 
     S & 6 & 64  & 0.75 &$64\times64\times3$  \\
     M & 5 & 1028 & 5.7 & $64\times64\times3$ \\ 
     L & 5 & 2048 & 6.3 & $64\times64\times3$  \\ 
\bottomrule
\end{tabular}}
\end{table*}

\section{Details on experiment setup}
\label{appdx:experiment_details}
In this section, we provide a detailed description on experiment setups for all experiment resuls presented in the paper. 

\paragraph{Dataset} \textbf{CIFAR-10} contains 50,000 training images from 10 classes, each with dimensions of $32\times 32 \times 3$.  \textbf{CIFAR-100} contains same number of images but more classes: 100 classes with 500 images each in the training data with dimension $32\times 32 \times 3$. \textbf{TinyImageNet} consists of 100,000 images distributed across 200 classes. The images within TinyImageNet are characterized by larger dimensions, measuring $64\times64 \times 3$. \textbf{ImageNet-1K} contains 1000 classes with around 1300 classes each, totalling 1.2 million images. We resized the images to $64\times64 \times 3$, aligning with prior works \citep{cui2022scaling, sachdeva2023data, zhou2023dataset}. Finally, we also include two known ImageNet subsets: \textbf{ImageNette} and \textbf{ImageWoof}. In line with established practices from prior work, we resize the images within both subsets of ImageNet to dimensions of $128\times 128 \times 3$. Each subset comprises 10 classes in their respective training sets, at total size of around 10k images.

\paragraph{Dataset preprocessing} For all three datasets, only a simple channel-based mean-variance scaling  is performed as the preprocessing step. For CIFAR-10 we perform ZCA whitening as done in all data distillation work \citep{nguyen2020dataset} using Kornia implementation with default parameters (\citep{riba2020kornia}. To generate experts used in MTT, we also perform random simple augmentations to the images, including rotations, flip, crop, and color changes. The preprocessing step is chosen to mirror the baselines we make direct comparisons to. 

\paragraph{Student network architecture} The student network is a neural network consists of multiple ConvNet blocks, and we call them ConvNet. The ConvNet configuration consists of multiple convolutional blocks, each housing a convolutional layer, a normalization layer, \texttt{ReLU} activation, and an average pooling layer. For larger datasets, we increase the number of convolutional blocks used in the ConvNet. For CIFAR10 we use ConvNet with 3 convolutional blocks, and for TinyImageNet and ImageNet-1K we use 4 convolution blocks. For ImageNet subsets, we use 5 convolutional blocks. In our MMD objective, we use the features generated by those convolutional blocks to compute MMD. Finally, a linear layer with Softmax activation is used to map the features generated by convolutional blocks into class prediction. 

\paragraph{Training} The distillation time is not the primary concern for data distillation tasks since it only needs to be done once for all downstream tasks. However, methods that are overly expensive to train might become infeasible when distilling large datasets. Because we compute and back propagate on both MTT and MMD losses, our compute time is comparable to both method combined. For CIFAR-10, CIFAR-100, ImageNette and ImageWoof, our method converges in fewer than 10,000 steps, usually taking less than 10 GPU hours on NVIDIA A100-SXM4-40GB. For TinyImageNet, our method converges around 10,000 steps, totalling around 20 GPU hours. Finally, for federated distilation on ImageNet, since we decompose into distillation sizes comparable to TinyImageNet, the training time is similar. 

For evaluation, we use SGD optimizer with momentum 0.9 and weight decay $5\times10^{-4}$. We only allow hyper-parameter tuning on the learning rate, number of epochs and we train student networks until convergence. 

\subsection{Decoder Hyperparameters}
\label{appdx:decoder_list}
In table \ref{tab:decoder_size_list}, we list the hyper-parameters we used for each setting and the exact storage costs to store the latent priors and decoders. 

\begin{table*}[h]
\centering
\caption{\textbf{Details on decoder hyper-parameters for all experiments} \textit{Decoder}: refer to Table \ref{tab:decoder_archi}, $\#$ \textit{Decoders}: more than one decoders for federated distillation when we aggregate from subtasks. \textit{LPC}: Latent Priors per Class refers to the number of Gaussian distributions we used to represent each class.}
\label{tab:decoder_size_list}
\resizebox{0.6\columnwidth}{!}{
\begin{tabular}{@{\extracolsep{-4pt}}c ccccccc}
\toprule   
    Dataset  & Decoder&  $\#$ Decoders & LPC  & Total Storage (MB)  \\
     \midrule
     CIFAR10 & S & 1& 10 & 3.4 \\
     CIFAR10 & S & 1& 50 & 7.8 \\
     CIFAR100 & S & 1& 2 & 4.2 \\
     CIFAR100 & M & 1 & 5 & 9.7 \\
     ImageNette & S & 1 &5 & 3.9 \\
     ImageWoof & S & 1 & 5 &3.9 \\
     TinyImageNet & S & 1  & 2 & 3.4 \\ 
     TinyImageNet & M & 1  & 2 & 10  \\
     TinyImageNet & L & 1 & 10 & 56 \\ 
     ImageNet-1K & S & 5 & 1& 17 &  \\ 
     ImageNet-1K & M & 2 & 2 & 76 &  \\ 
     ImageNet-1K & L & 5 & 10 & 440 &  \\ 
\bottomrule
\end{tabular}}
\end{table*}

\section{Additional Results}
\label{appdx:add_results}
Here we show additional results on CIFAR-100, CIFAR-10, ImageNette and ImageWoof, and compare our methods to methods that distill into image space: MTT \citep{cazenavette2022dataset} and FTD \citep{du2022minimizing} as well as methods that distill into latent space HaBa \citep{liu2022datset}, LinBa \citep{deng2022remember}  and GLaD \citep{cazenavette2023generalizing}. We report CIFAR-100 results in Table \ref{tab:cifar100}, CIFAR-10 results in Table \ref{tab:cifar10}, and ImageNette and ImageWoof results in Table \ref{tab:convnet_results_latent}. Additionally, we also report cross-architecture results for CIFAR-10 (see Table \ref{tab:cifar_cross}) and for ImageNette and ImageWoof (See Table \ref{tab:imagenet_cross_glad}, \ref{tab:imagenet_cross_haba}).

\begin{table}[h]
\centering
\caption{\textbf{CIFAR-100 distilled and evaluated on ConvNet}}
\label{tab:cifar100}
  \resizebox{0.5\columnwidth}{!}{
\begin{tabular}{@{\extracolsep{-10pt}}c cccccc}
\toprule   
    Storage Cost (MB)  & MTT & FTD & LinBa & D3(Ours)      \\
     \midrule
     5 \tiny{($\sim1$IPC)} &  \errbar{24.3}{0.3}  &   \errbar{25.2}{0.2}   & \errbar{34.0}{0.4} & \textbf{\errbar{37.3}{0.7}}  \\
     10  \tiny{($\sim10$IPC)} &  \errbar{40.1}{0.4} &  \errbar{43.4}{0.3}&    & \textbf{\errbar{46.8}{0.4}} &   \\
     50  \tiny{($\sim50$IPC)} &  \errbar{47.7}{0.3} & \errbar{50.7}{0.2}  &  &  & \\
\bottomrule
\end{tabular}}
\end{table}

\begin{table*}[h]
\centering
\caption{\textbf{CIFAR-10 distilled and evaluated on ConvNet}}
\label{tab:cifar10}
  \resizebox{0.5\columnwidth}{!}{
\begin{tabular}{@{\extracolsep{-4pt}}c cccccc}
\toprule   
    Storage Cost (MB)  & MTT & HaBa & LinBa & Ours      \\
     \midrule
     0.5 \tiny{($\sim$1IPC)} &  \errbar{46.3}{0.8}  &  \errbar{48.3}{0.8}    &  \errbar{66.4}{0.4} \\
     5 \tiny{($\sim$10IPC)} &  \errbar{65.3}{0.7} & \errbar{69.9}{0.4}   &    \errbar{72.2}{0.4} & \textbf{\errbar{71.8}{0.2}}  \\
     25  \tiny{($\sim$50IPC)} &  \errbar{71.6}{0.2} & \errbar{74.0}{0.2}  &   \errbar{73.6}{0.5} & \textbf{\errbar{74.4}{0.3}} \\
\bottomrule
\end{tabular}}
\end{table*}

\begin{table}[h]
 \caption{\textbf{ImageNette and ImageWoof distilled and evaluated on ConvNet} N/A: indicates the distillation size is smaller than the minimum size the method can distill.}
    \centering
      \resizebox{0.7\columnwidth}{!}{

    \begin{tabular}{@{\extracolsep{-4pt}}c c cccc}
    \toprule   
        \textbf{Dataset} & Storage (MB) &\multicolumn{4}{c}{\textbf{Method}} \\
        \midrule
         {}&  {}  & MTT & HaBa & FTD & Ours  \\     
         \cmidrule(lr){3-6}
        \textbf{ImageNette} & 5\tiny($\sim$0.5IPC) & N/A &  & N/A & \textbf{\errbar{71.04}{0.71}}\\
         {} & 10\tiny($\sim$1IPC) & \errbar{47.7}{0.9} & \errbar{51.92}{1.65} & \errbar{52.2}{1.0} &   \\
         {} & 100\tiny($\sim$10IPC) & \errbar{63.0}{1.3} & \errbar{64.72}{1.60} & \errbar{67.7}{0.7} &   \\
    
         \textbf{ImageWoof} &  5\tiny(0.5IPC) &  N/A  &  & N/A  & \textbf{\errbar{41.60}{1.15}} \\
         {} & 10\tiny($\sim$1IPC) & \errbar{28.6}{0.8} & \errbar{32.40}{0.67} & \errbar{35.8}{1.8} &   \\
         {} & 100\tiny($\sim$10IPC)  & \errbar{35.8}{1.8} & \errbar{38.60}{1.26} & \errbar{38.8}{1.4} &   \\
              
    \bottomrule
    \end{tabular}}
    \label{tab:convnet_results_latent}
    \end{table}

\begin{table}[h]
\caption{\textbf{CIFAR-10 cross-architecture generalization results} we scaled down our distilled dataset by reducing number of latent priors per class such that our performance on ConvNet aligns with baseline (MTT)}
    \centering
          \resizebox{0.7\columnwidth}{!}{

        \begin{tabular}{@{\extracolsep{-4pt}}ccccccc}
        \toprule   
        {} & & \multicolumn{4}{c}{\textbf{Evaluation Model}} \\
         \cmidrule(lr){3-6} 
         \textbf{Method} & Storage Cost (MB) & ConvNet & ResNet18 & VGG11& AlexNet \\ 
         \midrule
              MTT&  5 \tiny($\sim$10IPC)& \errbar{65.3}{0.7} & \errbar{46.4}{0.6} & \errbar{50.3}{0.8} & \errbar{34.2}{2.6} \\
             D3(ours)& 3 \tiny($\sim$10IPC)& \bf{\errbar{66.64}{0.26}} & \bf{\errbar{61.57}{0.48}} & \bf{\errbar{59.70}{0.48}} & \bf{\errbar{54.56}{0.74}} \\
        \bottomrule
        \end{tabular}}
    \vspace{5px}
    \label{tab:cifar_cross}
\end{table}

\begin{table}
    \centering
    \caption{\textbf{ImageNette and ImageWoof cross-architecture generalization results} Unseen architecture results from averaging ResNet18, VGG11, AlexNet, Vision Transformer.}
    \label{tab:imagenet_cross_glad}
      \resizebox{0.7\columnwidth}{!}{

    \begin{tabular}{@{\extracolsep{-8pt}}ccccccc}
    \toprule
    {} &  & \multicolumn{2}{c}{\textbf{ImageNette}} & \multicolumn{2}{c}{\textbf{ImageWoof}} \\
     \cmidrule(lr){3-4} 
    \cmidrule(lr){5-6} 
     \textbf{Method}  & Storage Cost (MB)  & ConvNet & Unseen & ConvNet & Unseen \\ \midrule
     MTT & 10\tiny{($\sim$1IPC)} & \errbar{47.9}{0.9} & \errbar{24.1}{1.8}&\errbar{28.6}{0.8}  & \errbar{16.0}{1.2}\\ 
     GLaD MTT& 10\tiny{($\sim$1IPC)} & \errbar{38.7}{1.6} & \errbar{30.4}{1.5} &\errbar{23.4}{1.1} &\errbar{17.1}{1.1}\\
     GLaD DC& 10\tiny{($\sim$1IPC)} &\errbar{35.4}{1.2} & \errbar{31.0}{1.6} & \errbar{22.3}{1.1}& \errbar{17.8}{1.1}\\
     GLaD DM& 10 \tiny{($\sim$1IPC)}& \errbar{32.3}{1.7}& \errbar{21.9}{1.1}  &  \errbar{21.1}{1.5}& \errbar{15.2}{0.9}\\ 
     D3(ours)& 5 \tiny{($\sim$0.5IPC)}&\textbf{\errbar{71.04}{0.7} } & \textbf{\errbar{48.95}{1.3}}& \textbf{\errbar{41.60}{1.2}}& \textbf{\errbar{28.82}{0.93}}\\
     \bottomrule
    \end{tabular}}
\vspace{5px}
\end{table}

\begin{table}[h]
\caption{\textbf{ImageNet Subset cross-architecture performances breakdown} Per-architecture breakdown for the unseen average listed in Table \ref{tab:imagenet_cross_glad}}
\label{tab:imagenet_cross_haba}
\centering
  \resizebox{0.8\columnwidth}{!}{
    \begin{tabular}{@{\extracolsep{-8pt}}ccccccc}
    \toprule{}
      {} & {} & \multicolumn{5}{c}{\textbf{Evaluation Model}} \\
      \cmidrule(lr){3-7}
      Dataset & Storage Cost (MB)  & ConvNet & ResNet & VGG11 & AlexNet & ViT \\ \midrule
       ImageNette & 5\tiny{($\sim$0.5IPC)} & \errbar{71.04}{0.71}  & \errbar{47.28}{0.94} &\errbar{64.80}{1.2} &\errbar{49.20}{1.45} & \errbar{34.5}{1.5} \\
      ImageWoof & 5\tiny{($\sim$0.5IPC)} & \errbar{41.60}{1.15} & \errbar{28.04}{0.51} & \errbar{35.48}{1.07}& \errbar{29.28}{1.4}& \errbar{22.48}{0.73}\\

     \bottomrule
    \end{tabular}}
\vspace{5px}
\end{table}
\clearpage
\section{Samples from distilled distribution}

\begin{figure}[hbtp]
    \centering
    \includegraphics[scale=0.08]{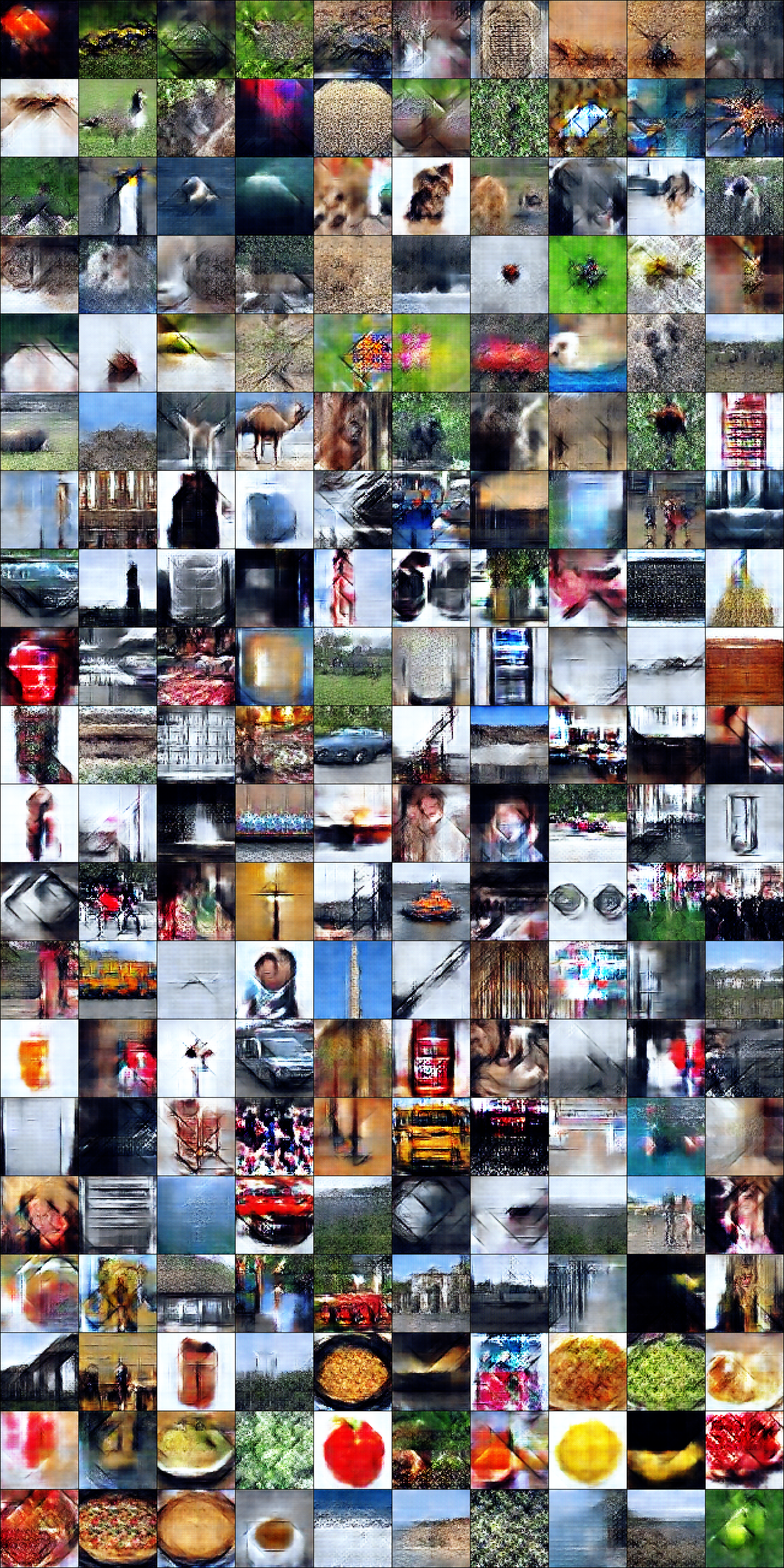}
    \caption{Samples from Distilled Distribution on TinyImageNet under 100MB storage cost.  1 latent priors per class visualized}
\end{figure}

\begin{figure}[ht]
    \centering
    \includegraphics[scale=0.7]{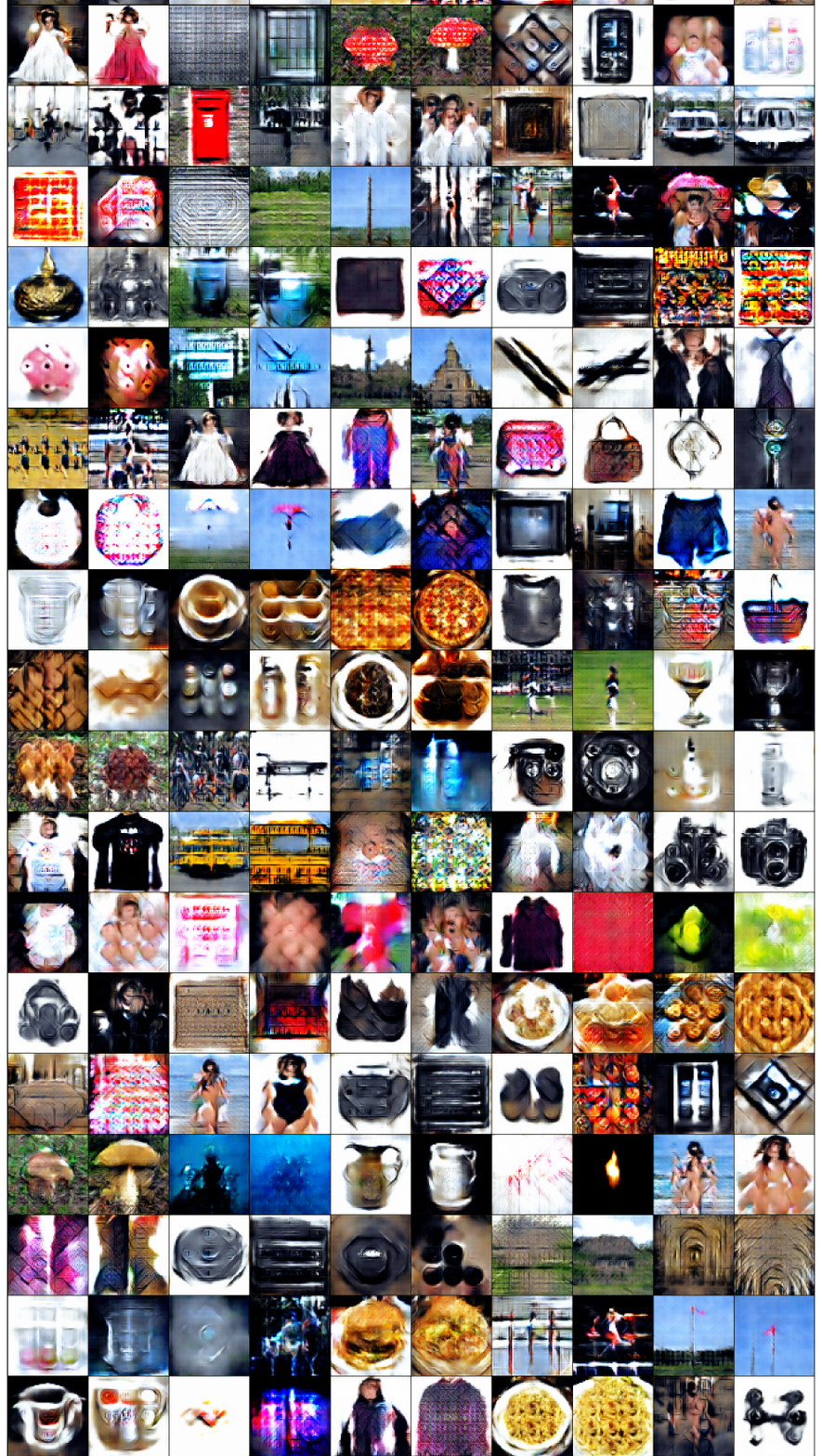}
    \caption{Samples from Distilled Distribution on ImageNet-1K under 100MB storage cost. 2 latent priors per class visualized}
\end{figure}

\begin{figure}[ht]
    \centering
    \includegraphics[scale=0.7]{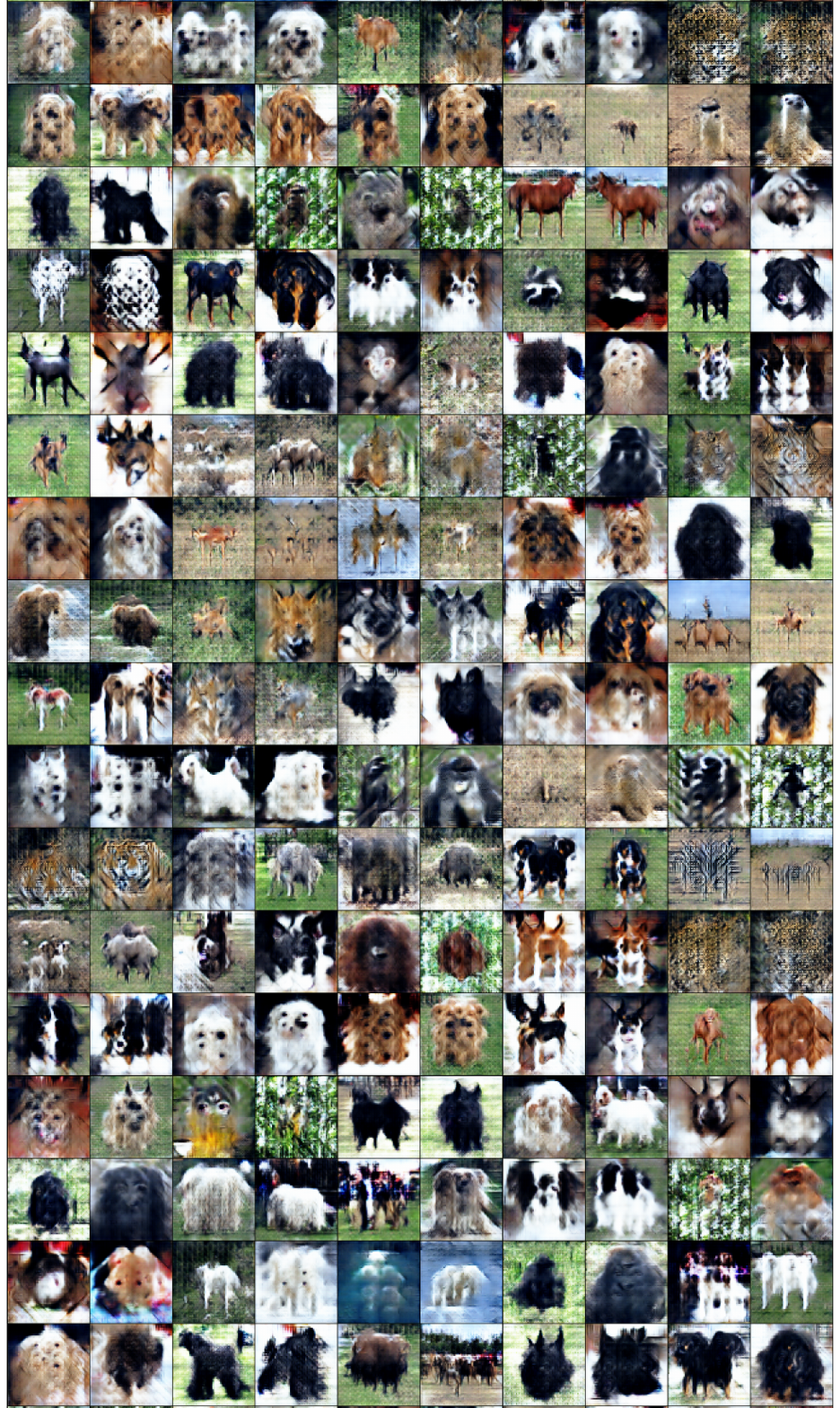}
    \caption{Samples from Distilled Distribution on ImageNet-1K under 100MB storage cost (continued).  2 latent priors per class visualized}
\end{figure}


\end{document}